\documentclass[10pt,twocolumn]{article}

\usepackage[utf8]{inputenc}
\usepackage[T1]{fontenc}
\usepackage{amsmath,amssymb,amsfonts,amsthm}
\usepackage{graphicx}
\usepackage{booktabs}
\usepackage{multirow}
\usepackage{array}
\usepackage{url}
\usepackage[hidelinks]{hyperref}
\usepackage{geometry}
\usepackage{caption}
\usepackage{subcaption}
\usepackage{enumitem}
\usepackage{xcolor}
\usepackage{algorithm}
\usepackage{algpseudocode}
\usepackage{balance}
\usepackage{pgfplots}

\newtheorem{proposition}{Proposition}
\newtheorem{lemma}{Lemma}
\newtheorem{remark}{Remark}
\newtheorem{definition}{Definition}
\newtheorem{corollary}{Corollary}

\DeclareMathOperator{\KL}{KL}
\DeclareMathOperator{\SNR}{SNR}

\makeatletter
\newcommand\blfootnote[1]{%
  \begingroup
  \renewcommand\thefootnote{}%
  \long\def\@makefntext##1{\noindent ##1}%
  \footnotetext{#1}%
  \addtocounter{footnote}{-1}%
  \endgroup
}
\makeatother

\title{\bfseries The Paired Thermal Channel Is Two Orthogonal Axes: A Controlled Attribution Study of Calibration Cost and Detection Gain in Infrared Person Detection}

\author{
Tian Zhang,
Yang Chen,
Bing Bai,
Weikai He,
Xiang Gao,
and Zhirong Su
}

\date{}

\begin{document}

\maketitle

\blfootnote{%
T. Zhang and B. Bai are with the School of Information Engineering, Shandong Management University, Jinan, China. \\
Y. Chen is with the School of Electrical and Electronic Engineering, Nanyang Technological University, Singapore (Corresponding author). \\
X. Gao and W. He are with the School of Aeronautics, Shandong Jiaotong University, Jinan, China.\\
Z. Su is with the School of Business, Shandong Management University, Jinan, China.\\
Email:\texttt{~tianzhang@sdnu.edu.cn, chen1605@e.ntu.edu.sg, bbingsdmu@126.com, 214035@sdjtu.edu.cn, abrahamgx@163.com, suzhirong007@163.com}%
}

\begin{abstract}
The calibrated thermal channel in infrared person detection comprises two structurally orthogonal axes. True calibrated $\Delta T$ outperforms scale-matched grayscale pseudo-$\Delta T$ by $+0.0221$ mAP (paired Wilcoxon $p=0.0098$, Cohen's $d=+1.08$), confirming that the calibrated channel carries information the grayscale channel lacks. The calibration axis is driven by the joint pairing-scale condition. Preserving both paired correspondence and absolute calibrated scale raises expected calibration error (ECE) from $\approx 0.05$ to $0.142$. The $2\times 2$ interaction ($p=0.0059$) localizes the rise to this condition. Breaking either factor alone returns ECE to the broken-factor level (true-vs-const $p=0.002$; true-vs-shuffled $p=0.002$). Under post-hoc temperature scaling, the pairing cost survives recalibration ($p=0.002$), while the scale cost is largely absorbed ($p=0.16$). As regards the detection axis, the gain stems from structural de-structuring. Phase randomization, which destroys spatial structure while preserving per-sample variance, yields $+0.0185$ mAP over true $\Delta T$ on FLIR ADAS ($p=0.014$, $d=+0.96$). The FLAME~3 gain is $+0.0064$ mAP ($p=0.078$, n=8). On the calibration axis, the same intervention is insignificant on FLIR ADAS ($p=0.47$) but significant on FLAME~3 ($p=0.0078$). These findings rest on a five-arm controlled attribution with temperature-evidential learning (TEL) as the testbed. TEL obtains a single evidence-derived routing scalar together with a complementary diagnostic entropy, and establishes a finite-evidence degeneracy theorem proving that the acquisition product degenerates to this scalar with a closed-form bound. A physical $0.5^\circ$C threshold admits cross-dataset physical readings, whereas a gray threshold of $20$ inverts without physical interpretation.
\end{abstract}


\section{Introduction}
\label{sec:intro}

Pedestrian detection at night or under other low-visibility conditions remains a persistent challenge for thermal infrared perception. Among the sensing modalities that stay reliable under low illumination, thermal infrared is distinguished by its stable response across lighting conditions. The literature has advanced from handcrafted features to end-to-end convolutional detectors, with recent work on multispectral pedestrian detection transformers~\cite{msdetr2024}, illumination-aware thermal detection~\cite{li2024illumination}, and thermal image enhancement~\cite{mudavath2026review}. However, the treatment of radiometric data has remained largely unchanged. A thermal camera outputs a calibrated temperature per pixel. Conventional pipelines convert this into an 8-bit grayscale image or a false-color rendering and then discard the temperature matrix, which carries the physical evidence. When a person wears heavy clothing or stands before a warm surface, the grayscale contrast collapses and the detector loses the target.

The issue intensifies within learning loops. Active learning chooses samples to annotate, pseudo-labeling chooses which predictions to trust, and continual learning chooses which past examples to replay. When only a rendered image is available, all such decisions rely on the same visual features and inherit the same blind spots. Evidential deep learning (EDL) offers a principled framework for uncertainty estimation. It places a Dirichlet prior over class probabilities and gathers evidence in a single forward pass~\cite{sensoy2018edl,gao2024survey}. EDL has been brought into detection~\cite{barbarit2026evyolo,zhao2026openworld,yang2026mmlf,li2024stabilizing}, and evidence-based fusion has been explored for multispectral pedestrian detection. Yet these methods still treat uncertainty as a purely visual quantity, and the applications of EDL to active learning and pseudo-labeling have progressed on separate tracks. Existing frameworks address annotation, pseudo-labeling, and replay with separate criteria.

We examine what $\Delta T$, i.e., the temperature contrast between a target region and its immediate surround, actually is. It is a calibrated physical measurement. Its physiological source is stable across scenes and sensors: human core temperature remains near 310~K, although clothing and background can attenuate the observed surface contrast. $\Delta T$ is also correlated with the local separability of the target from its background. Among the visual features considered in this work, these three properties are unique to the calibrated thermal channel. These properties indicate that $\Delta T$ is a reliable signal for evidence modulation. The resulting uncertainty can then drive annotation, pseudo-labeling, and replay.
\begin{figure*}[!t]
\centering
\includegraphics[width=\textwidth]{"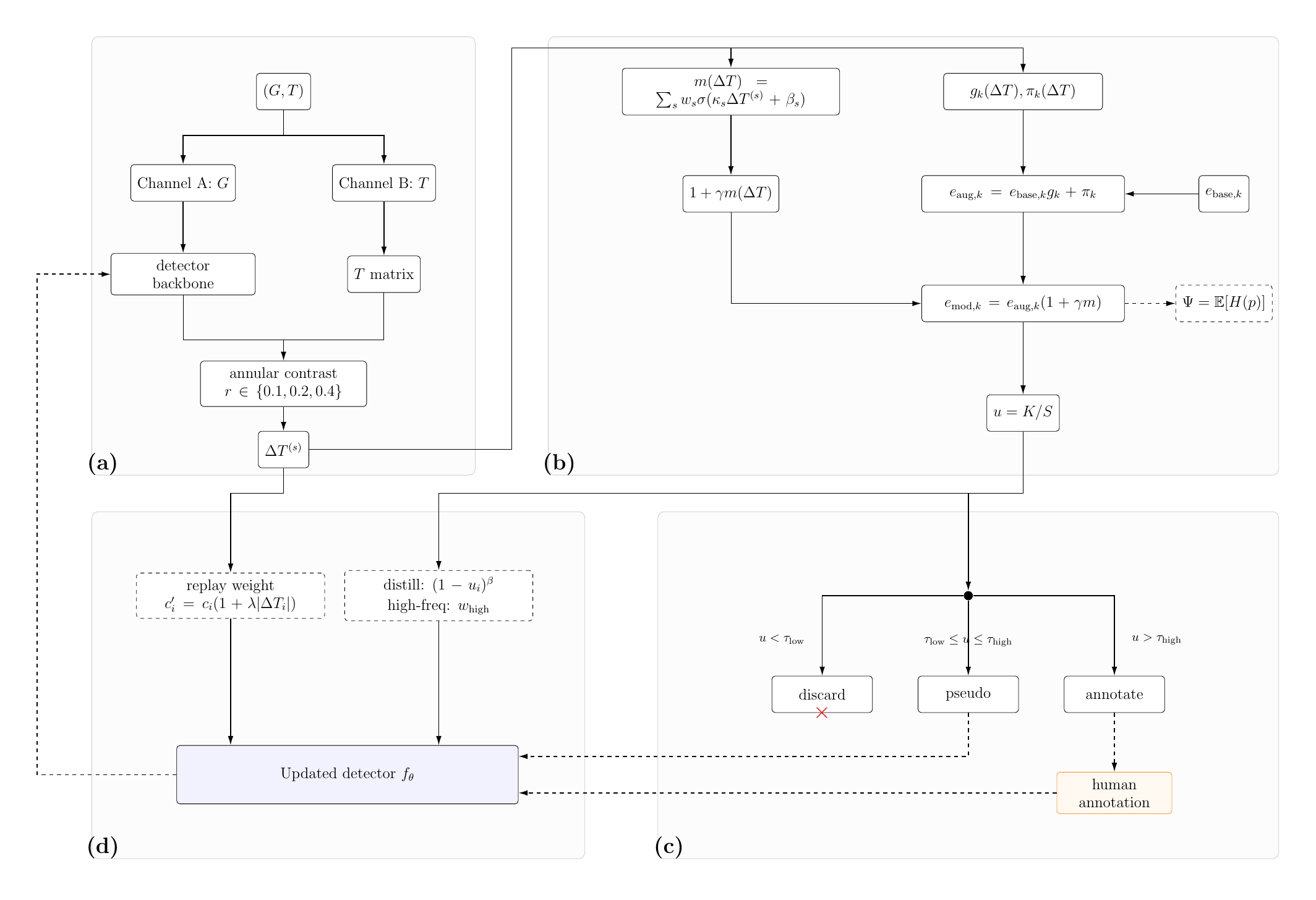"}
\caption{Temperature-evidential learning (TEL) as one evidence chain.
(a)~Dual-channel decoding. Channel~A renders $G$, while Channel~B retains the calibrated temperature matrix $T$.
The annular contrast geometry is sampled at dilation factors $r \in \{0.1, 0.2, 0.4\}$,
using $T$ values from Channel~B and box positions from the detector backbone.
(b)~Temperature-evidential modulation. The multi-scale contrast $\Delta T^{(s)}$
(Eq.~\eqref{eq:multiscale}) passes through the sigmoid gate $m(\Delta T)$
(Eq.~\eqref{eq:multiscale_mod}). The temperature-aware prior $g_k, \pi_k$
produces $e_{\text{aug},k}$ (Eq.~\eqref{eq:temp_prior}), and the modulation
factor $1 + \gamma m$ produces $e_{\text{mod},k}$ (Eq.~\eqref{eq:evidence}),
from which a single evidence-derived routing scalar $u$ is read; a diagnostic entropy $\Psi=\mathbb{E}[H(p)]$ is computed alongside on a dashed branch and does not participate in routing.
(c)~Two-ended routing. The same $u$ drives annotation, pseudo-labeling, and
discard (Eq.~\eqref{eq:routing}); the partition relies on separation of the two ends of the distribution, not on monotone ranking in between.
(d)~Replay and distillation weighting (Eqs.~\eqref{eq:tempweight} and~\eqref{eq:distillloss})
appear as dashed design extensions.
Solid arrows denote forward data flow and core computation. Dashed arrows denote auxiliary inputs, control signals, and the closed-loop feedback that returns the updated detector to the next round. All downstream decisions in (b)-(d) use the temperature matrix retained in Channel~B.}
\label{fig:lifecycle}
\end{figure*}

This paper presents a structural attribution finding for paired physical channels. The calibrated thermal channel has two components with different causal origins. ECE responds to the joint pairing-scale condition, where the pairing of target and background and the absolute calibrated scale act jointly on it. mAP responds to a different factor, namely how the thermal contrast is injected into the modulation channel, not which spatial thermal pattern the contrast preserves. We establish these two axes through five controlled attribution arms. The true arm retains all five factors. The other three arms each break one of them. Specifically, const-$\Delta T$ keeps within-image structure but drops the absolute calibrated scale, shuffled-$\Delta T$ keeps the amplitude but destroys paired correspondence, and var-phase keeps both the power spectrum and the per-sample variance while removing only the spatial phase. The random arm breaks all four factors (pairing, scale, power-spectrum structure, and spatial phase) at once, with the injected variance held matched. The five arms separate the five factors and expose the two axes. These axes are instantiated by the evidence chain in Fig.~\ref{fig:lifecycle}. Retaining the calibrated matrix $T$ enables modulation, which anchors a single cognitive uncertainty $u$ as the routing scalar. TEL implements this chain and serves as the attribution testbed.

The contributions consist of attribution findings and the methodological framework that supports them.
\begin{itemize}[leftmargin=1.2em, itemsep=1pt]
\item \textbf{Pairing-scale joint condition as an operational criterion.} A 2$\times$2 intervention (true, const-$\Delta T$, shuffled-$\Delta T$, and random) separates paired correspondence from absolute calibrated scale, and the interaction contrast tests whether the two factors act jointly. Preserving both raises ECE from $\approx 0.05$ to $0.142$ (interaction Wilcoxon $p=0.0059$). Breaking either factor in isolation returns it to the broken-factor level (true-vs-const $p=0.002$ and true-vs-shuffled $p=0.002$). Post-hoc temperature scaling separates the two factors by how much of their ECE cost a monotone rescaling can remove. The true-vs-shuffled gap survives recalibration ($p=0.002$), whereas the true-vs-const gap is absorbed ($p=0.16$). The two factors act on the same confidence distribution, and the distinction is one of proportion, not of kind. The criterion is operational, as any calibrated physical channel can be tested with the same four-arm intervention plus a post-hoc recalibration probe, and we validate it on two radiometric benchmarks.

\item \textbf{The calibrated channel carries information.} True $\Delta T$ outperforms scale-matched grayscale pseudo-$\Delta T$ by $+0.0221$ mAP (paired Wilcoxon $p=0.0098$, $d=+1.08$), so the calibrated channel carries information the collapsed grayscale channel does not.

\item \textbf{Detection gain tracks structural de-structuring, not spatial thermal structure.} Random modulation exceeds true $\Delta T$ by $+0.0341$ mAP, and the phase-randomized arm exceeds true $\Delta T$ by $+0.0185$ mAP (paired Wilcoxon $p=0.014$, $d=+0.96$) while preserving per-sample variance and removing only spatial phase. Flattening the power spectrum accounts for the remaining $+0.0156$ ($p=0.004$). The gain is attributable to the destruction of spatial structure, not to the spatial thermal pattern that calibration preserves. The two axes vary independently. In the cross-dataset contrast, each axis is significant on the dataset where its structural counterpart is most pronounced.

\item \textbf{Deployment value of the retained absolute scale.} Physical thresholds expressed in $^\circ$C remain semantically consistent across datasets. By contrast, a gray threshold of $20$ inverts across the same pair and admits no physical reading.

\item \textbf{A modality-agnostic attribution protocol.} The five-arm design, 2$\times$2 interaction Wilcoxon, and partial-correlation control are defined on a generic calibrated auxiliary channel rather than on temperature specifically. We validate the protocol on two radiometric benchmarks. These two benchmarks span two orders of magnitude in $|\Delta T|$ and differ in the $\Delta G$--$\Delta T$ Pearson correlation, so the protocol is tested under two structurally different calibrated regimes. The same construction applies to any paired channel that provides a calibrated auxiliary measurement alongside a visual one, such as LiDAR reflectance or radar cross-section.

\item \textbf{A unified routing framework with proved consistency.} A single scalar $u$ governs annotation, pseudo-labeling, and discard through one rule. The routing partition is invariant under strictly increasing affine reparametrizations, and the acquisition product degenerates to $u$ in the finite-evidence regime ($R^2 \geq 1-\eta$). We establish the framework and prove it self-consistent.
\end{itemize}

\section{Related Work}
\label{sec:related}

\subsection{Uncertainty Estimation in Deep Learning}
\label{sec:related_uq}

Bayesian neural networks and deep ensembles provide principled uncertainty estimates but are expensive at inference. Monte Carlo dropout approximates the posterior by treating test-time dropout as variational inference~\cite{gal2016dropout}, while deep ensembles require training several independent networks~\cite{lakshminarayanan2017ensembles}. EDL, introduced by Sensoy et al.~\cite{sensoy2018edl}, avoids both costs. The network outputs the parameters of a Dirichlet distribution over class probabilities, and the total evidence determines both the per-class belief mass and the residual uncertainty. Gao et al.~\cite{gao2024survey} survey the theoretical foundations and a wide range of downstream applications.

In detection, Ev-YOLO~\cite{barbarit2026evyolo} unifies classification and localization uncertainty under a single evidential formulation. It finds that localization uncertainty separates correct detections from erroneous ones under domain shift on KITTI, MUSES, and nuScenes. Zhao et al.~\cite{zhao2026openworld} use evidential learning for open-world detection. EviRCOD~\cite{evircod2026} introduces evidence-guided decoding for camouflaged object detection, and EUGNet~\cite{eugnet2026} applies a similar idea to underwater salient object detection. Recent work has also explored evidential regression with Normal-Inverse Gamma distributions for 3D detection~\cite{zhao2025evmvx}, and deterministic uncertainty quantification via hyperdimensional computing~\cite{chen2025hyperdimensional}.

Several studies have identified systematic limitations of EDL. Bengs et al.~\cite{bengs2022difficulty} argue that EDL-style methods are highly sensitive to the regularization hyperparameter and tend to underestimate epistemic uncertainty. Re-EDL~\cite{reedl2024} provides evidence that the KL regularization term can itself exacerbate overconfidence. Ulmer et al.~\cite{ulmer2023prior} question the calibration reliability of evidential uncertainty under distribution shift. These critiques imply that a high ECE in an EDL detector may be an artifact of the evidential loss, not a property of the input channel. In TEL, all attribution arms---true, pseudo, random, shuffled, const-$\Delta T$, and var-phase---share the same EDL loss, the same regularization weight, and the same modulation architecture. Any KL-regularization effect is therefore common across arms and cancels in the paired contrasts. This motivates treating the physical channel as an evidence variable, not as a preprocessing input. TEL feeds the temperature contrast directly into the Dirichlet prior, so that the joint pairing-scale condition enters the evidence model.
\subsection{Confidence Calibration in Detection}
\label{sec:calibration_related}

ECE is the primary arbiter of the calibration-cost claim, so we review the calibration literature separately. Guo et al.~\cite{guo2017calibration} introduced the modern ECE estimator and concluded that deep networks are systematically overconfident. Temperature scaling was introduced as a simple post-hoc remedy. This estimator underlies the ECE used throughout this paper, and Section~\ref{sec:step_which} applies it to measure how much of each factor's ECE cost a monotone rescaling can absorb.

In object detection, calibration is more subtle than in classification because confidence is entangled with localization quality and background imbalance. Cal-DETR~\cite{caldetr2023} introduced train-time calibration for detection transformers and proposed a detection-aware ECE (D-ECE) that accounts for localization. Kuzucu et al.~\cite{kulin2024calibration} provide a systematic study of detector calibration pitfalls, evaluation variants, and baselines, and find that post-hoc calibration methods, when designed properly, can outperform train-time calibration.

Calibration under domain shift and in safety-critical perception has also been studied. Cal-DETR reports out-of-distribution calibration behaviour, and uncertainty-aware LiDAR/camera fusion work uses aleatoric calibration metrics such as ACE. These works motivate our emphasis on calibration as a deployment concern. They operate on visual or LiDAR features, not on a calibrated physical measurement channel. TEL instead operates on a calibrated physical measurement channel; the calibration cost then appears as a joint pairing-scale condition.

\subsection{Active Learning and Pseudo-Labeling}
\label{sec:related_al}

Active learning reduces annotation cost by querying the most informative unlabeled samples, and pseudo-labeling expands the training set with model predictions. The deep active learning literature has moved from heuristic acquisition functions to representation-based and uncertainty-aware strategies~\cite{ren2021survey}. Li et al.~\cite{li2025survey} cover batch-mode selection, deep Bayesian active learning, and hybrids that mix uncertainty with diversity. Pseudo-labeling assigns labels when confidence exceeds a threshold. Modern frameworks combine the two: pseudo-labeling handles the bulk of high-confidence data, while active learning targets the uncertain remainder~\cite{medicalsurvey2024}. The difficulty with this combination is that annotation and pseudo-labeling typically rely on separate scoring functions, which can lead to inconsistent decisions across rounds and confirmation bias when pseudo-labels are generated from overconfident predictions~\cite{arazo2020pseudo}.

Evidential uncertainty has begun to be used to address this issue. CSEAL integrates EDL into semi-supervised active learning and leverages evidential uncertainty to prioritize annotation~\cite{balaram2022cseal}, but it still treats uncertainty as a selection score as opposed to a routing criterion that also governs pseudo-labeling and retention. EDALSSC~\cite{zhao2025edalssc} quantifies uncertainty for labeled and unlabeled data and selects samples by a sum-based score when the training loss rises late in learning. FEAL~\cite{feal2024} introduces Dirichlet priors in both local and global models for calibrated evaluation under domain shift. Evidential active learning with determinantal point processes~\cite{dpp2025} considers uncertainty and diversity together for batch selection. In detection, MGCAMT~\cite{mgcam2025} models category uncertainty via evidential deep learning and filters out incorrect pseudo-labels through an uncertainty-aware selection strategy. Adaptive frameworks~\cite{adaptiveal2026} balance uncertainty and diversity, and evidential hierarchical aggregation has been applied to detection as well~\cite{evidalod2026}. This line of work uses separate criteria for annotation, pseudo-labeling, and discard. TEL unifies them through a single temperature-modulated scalar.

\subsection{Continual Learning and Replay}
\label{sec:related_cl}

Catastrophic forgetting is a central challenge in continual learning, and replay is among the most effective methods to counter it. The critical question is which historical samples to store. GEM and its averaged variant project gradients to prevent interference but do not select at the sample level. GMIR~\cite{nisar2023gmir} retrieves samples whose gradients maximally interfere with the current update, reducing training time on SeeingThroughFog by 46.25\%. Balanced retrieval of conflicting and aligned samples has also been explored~\cite{xu2024balanced}.

GDA~\cite{luo2025gda} uses a Gaussian Mixture Model to estimate pseudo-labels of previously learned objects in current images and then aligns gradients to maintain stability while allowing positive transfer. FAS~\cite{fas2026} focuses on discriminative feature learning, aligns query-target assignments, and sustains feature-space support for old classes through manifold-support replay. RePRE~\cite{wu2025repre} constructs coarse and fine-grained prototypes from RoI features and observes that forgetting in two-stage detectors is concentrated in the RoI Head classifier and not the regressor. YOLO-IOD~\cite{yolo_iod2026} extends incremental object detection to a real-time YOLO framework and identifies three knowledge-conflict categories that must be addressed jointly. These methods base the selection decision on visual or gradient-based criteria. TEL weights replay by $|\Delta T|$ and by the temperature-modulated uncertainty, so the memory and transfer inherit the evidence ranking.

\subsection{Infrared Pedestrian Detection and Cross-Modal Distillation}
\label{sec:related_ir}

Infrared pedestrian detection has advanced considerably. Vachhani et al.~\cite{vachhani2024lowlight} survey low-light pedestrian detection in visible and infrared image feeds. Recent methods include HMFIDet~\cite{hmfidet2025} with hierarchical feature fusion, TFNet~\cite{chu2024toward} with thermal-first and fusion-second design, and MS-DETR~\cite{msdetr2024} with loosely coupled fusion and modality-balanced optimization. Strip-Fusion~\cite{stripfusion2026} handles misalignment and lighting variation with temporally adaptive convolutions and a KL-divergence loss. DAMSDet~\cite{damsdet2024} introduces a dynamic adaptive multispectral detection transformer with competitive query selection and adaptive feature fusion, addressing modality complementarity and misalignment jointly. MGIR-Net~\cite{MIGRNET2026} is designed for visible–infrared object detection in UAV remote sensing.

Infrared datasets are small relative to visible-light ones, and cross-modal distillation from visible-light teachers is one approach to this data-scale constraint. Naive alignment across bands causes negative transfer, however, because the two modalities diverge spectrally. FreqKD~\cite{thaker2026freqkd} observes that high-frequency cross-modal differences are 2.4$\times$ larger than low-frequency ones, and proposes asymmetric supervision, with strict MSE on low frequencies to preserve structure and relaxed log-MSE on high frequencies to provide edge guidance without forcing texture agreement. FreqKD reaches 64.1 mAP@0.5 on KAIST, 2.4 points above its DINOv2 baseline, with positive transfer to FLIR ADAS~\cite{flir_dataset} and MFNet. Thermal-Det~\cite{thermaldet2026} synthesizes million-scale thermal data from RGB and distills open-vocabulary capability from a frozen RGB detector. KDDFuse~\cite{kddfuse2026} fuses infrared and visible via frequency-decoupled distillation and local structure alignment. All of these methods use globally fixed band weights, whereas TEL makes the weight a function of the physical contrast $\Delta T$, so the supervision strength adapts per sample.

\subsection{Temperature Information in Radiometric Imagery}
\label{sec:related_temp}

The temperature matrix in radiometric thermal images is a calibrated physical field. Son et al.~\cite{son2017radiometric} built a radiometric dataset and a TIR-ACF feature, improving detection by 5\%. Okuda et al.~\cite{okuda2024realtime} studied ambient-temperature effects on multispectral detection, and noted that a model trained on a single temperature subset degrades on others while a mixed-temperature training set recovers accuracy. Lu et al.~\cite{lu2024combining} combined gray stretch with temperature for deep learning preprocessing. Thermal victim detection in collapsed buildings~\cite{thermalvictim2026} uses posture and temperature for rescue prioritization. Across these works, temperature enters at the preprocessing or feature-extraction stage. TEL uses it as a structural variable in uncertainty modeling. The calibration cost then follows the joint pairing-scale condition.

\subsection{Multi-Modal Uncertainty Fusion and Modality Reliability}
\label{sec:multimodal_reliability}

Uncertainty-aware fusion across sensor modalities has become an active area for autonomous driving. Yang et al.~\cite{yang2026mmlf} develop a decision-level fusion based on Dempster-Shafer theory and obtain substantial uncertainty reduction for cars, pedestrians, and cyclists on KITTI. UGFusion~\cite{ugfusion2026} propagates uncertainty through the entire bird's-eye-view pipeline and introduces an uncertainty-align fuser. UDF-3D~\cite{hu2026udf} addresses decision-level fusion for camera-LiDAR 3D detection, noting that most methods rely on object confidence and ignore object uncertainty. Uncertainty-aware federated learning for multi-sensor fusion~\cite{alonso2026federated} separates aleatoric and epistemic uncertainty in vehicular perception. These methods operate on visual or LiDAR features.

A closely related problem is weak spatial alignment between modalities. Zhang et al.~\cite{zhang2019weakly} introduced weakly aligned cross-modal learning for multispectral pedestrian detection, and Chen et al.~\cite{chen2024weakly} proposed misalignment-free adaptive feature alignment for UAV-based multimodal object detection. Strip-Fusion~\cite{stripfusion2026} explicitly models misalignment with temporally adaptive convolutions. Modality-reliability modeling forms another neighbouring line. Yang et al.~\cite{yang2026mmlf} down-weight unreliable modalities at the decision level, and Li et al.~\cite{li2024stabilizing} use Dirichlet evidence and Dempster--Shafer fusion to stabilize multispectral pedestrian detection under modality corruption. These works establish paired correspondence and modality reliability as first-class variables in multimodal detection. By comparison, TEL keeps the calibrated temperature matrix and treats the target-background contrast as the evidence-modulation channel, going beyond test-time reweighting. Its shuffled-$\Delta T$ probe intervenes directly on paired correspondence while preserving amplitude, which lets the joint pairing-scale condition be tested as an attribution question as opposed to only a robustness benchmark.

\emph{Summary of the Gap:} EDL, active learning, pseudo-labeling, replay, frequency-decoupled distillation, detector calibration, and modality-reliability fusion are existing techniques that have been combined previously. What remains unclear is how they should be combined. To the best of our knowledge, no prior framework treats a calibrated physical channel as the evidence variable that determines how annotation, pseudo-labeling, replay, and distillation should be combined. In existing pipelines, those decisions still use separate scores, and temperature remains a preprocessing or feature-extraction quantity but not a structural variable in uncertainty modeling. The detection-calibration literature has documented that ECE is definition-sensitive and that post-hoc calibration can be strong. TEL instead examines whether a calibrated physical input channel changes the intrinsic confidence distribution of an evidential detector. The misalignment and modality-reliability literature has studied pairing and reliability as robustness variables. TEL treats them as attribution variables for calibration cost.

\section{Method}
\label{sec:method}

\subsection{Problem Setup}
\label{sec:problem}

A radiometric thermal image is a pair $(G, T)$, where $G \in \mathbb{R}^{H \times W}$ is a grayscale rendering and $T \in \mathbb{R}^{H \times W}$ is the calibrated temperature matrix. The label set is $\mathcal{Y} = \{1,\dots,K\}$ with background included, and a detector $f_\theta$ proposes boxes $\mathcal{B}(x) = \{b_1,\dots,b_M\}$ with class predictions and evidence parameters. We aim to train $f_\theta$ under a fixed annotation budget $\mathcal{C}$ while retaining performance across a sequence of domains $\{\mathcal{D}_1,\dots,\mathcal{D}_N\}$. In this setting, the annotation budget is allocated among detection, pseudo-labeling, and replay decisions.

Within this setting, the central question is \emph{where the calibration cost comes from}. This cost arises from the joint pairing-scale condition, established by two stages. The first stage, presented in this section, confirms that the temperature channel carries separable information and that this information uniformly drives the data lifecycle. The second stage, developed in Section~\ref{sec:experiments}, decomposes the calibration cost into its pairing and scale components through controlled attribution probes. A parallel decomposition of the detection axis into its structural and variance components follows in the same attribution chain.

To allocate the annotation budget within this general setting, we solve
\begin{equation}
\begin{aligned}
\lefteqn{\min_{\theta, \pi} \quad
 \mathbb{E}_{x \sim \cup_{n=1}^{N} \mathcal{D}_n}\!\Bigl[
    \mathcal{L}_{\text{det}}(f_\theta(x), y)
    + \lambda_1\, \pi(x)}\\
& \qquad
   + \lambda_2\, \mathcal{L}_{\text{replay}}(x; c'(x)) + \lambda_3\, (1-u(x))\, \mathcal{L}_{\text{distill}}(x) \Bigr] \\
& \text{s.t.} \quad
\sum_{n=1}^{N} \lvert \mathcal{A}_n \rvert \le \mathcal{C},
    \quad \pi(x) \in \{0, 1\},
\end{aligned}
\label{eq:unified}
\end{equation}
where $\pi(x)$ indicates whether $x$ is annotated, $\mathcal{A}_n$ is the annotated set in round $n$, and the four terms correspond to detection, fixed annotation cost, replay, and distillation. The annotation cost is independent of sample uncertainty, as routing rule Eq.~\eqref{eq:routing} selects high-information-gain samples via $u$, while the budget constraint $\sum_n |\mathcal{A}_n| \le \mathcal{C}$ limits the total number of annotations. Proposition~\ref{prop:routing_bound}(i) establishes that \(u(x)\) is monotone in the leading term of the expected information gain in the large-evidence limit. Under the fixed budget, this work ranks by \(u\) and annotates the top samples.

The first stage decomposes into four layers. The physical layer concerns the separability of the paired thermal channel. The information-theoretic layer explains the KL separability of the temperature channel. The decision layer covers unified routing via cognitive uncertainty, and the optimization layer treats closed-loop convergence. Fig.~\ref{fig:lifecycle} summarizes the chain.

\subsection{Physical Layer: Separability of the Paired Thermal Channel}
\label{sec:tempevidential}
For a candidate box $b$, let $R_{\text{obj}}(b)$ be the interior and let
\begin{equation}
R_{\text{bg}}(b) = \mathcal{E}_r(b) \setminus R_{\text{obj}}(b)
\label{eq:annulus}
\end{equation}
be an annulus obtained by dilating the box with a factor $r$ and removing the interior. For a box $b$ with width $w$ and height $h$ centered at $(x_c, y_c)$, the dilated box $\mathcal{E}_r(b)$ has dimensions $(1+r)w \times (1+r)h$ with the same center. We use $r \in \{0.1, 0.2, 0.4\}$, corresponding to annular widths of $5\%$, $10\%$, and $20\%$ of the box dimension. A global background statistic would mix the target's thermal context with unrelated surfaces. The annulus keeps the reference local. The dual-channel decoding and annular geometry are summarized in Fig.~\ref{fig:lifecycle}(a,b).

A single annular scale may not capture the full thermal context. A small annulus emphasizes sharp boundaries but is sensitive to noise, while a large annulus gives a robust background statistic but may include unrelated surfaces. We therefore compute a multi-scale contrast
\begin{equation}
\begin{aligned}
\Delta T^{(s)}(b) =\;& \frac{1}{\lvert R_{\text{obj}}(b) \rvert}\sum_{i \in R_{\text{obj}}(b)} T_i \\
&-\; \frac{1}{\lvert R_{\text{bg}}^{(s)}(b) \rvert}\sum_{j \in R_{\text{bg}}^{(s)}(b)} T_j,
\end{aligned}
\label{eq:multiscale}
\end{equation}
where $R_{\text{bg}}^{(s)}(b)$ is the annulus at dilation factor $r_s$, for $s = 1,\dots,S$ with $r_s \in \{0.1, 0.2, 0.4\}$ in our experiments. When $S=1$, Eq.~\eqref{eq:multiscale} reduces to the single-scale contrast. The sign is retained, with $\Delta T > 0$ for warm-target/cool-background and $\Delta T < 0$ for shade on a sun-heated surface. Both polarities are passed to the modulation. No absolute value is taken, so the detector can distinguish the two physical regimes.

A learned attention module combines the scales:
\begin{equation}
\begin{aligned}
m(\Delta T) &= \sum_{s=1}^{S} w_s \cdot \sigma(\kappa_s \Delta T^{(s)} + \beta_s), \\
w_s &= \frac{\exp(\psi_s(\Delta T^{(s)}))}{\sum_{s'=1}^{S} \exp(\psi_{s'}(\Delta T^{(s')}))},
\end{aligned}
\label{eq:multiscale_mod}
\end{equation}
where $\psi_s$ is a two-layer  multi-layer perceptron (MLP) with hidden dimension 16 and $\kappa_s > 0$ are learnable slope parameters. The weights sum to one and are data-dependent. When $S = 1$, Eq.~\eqref{eq:multiscale_mod} reduces to the single-scale sigmoid modulation with slope $\kappa_1$.

Before using $m$ inside the Dirichlet prior, we justify the sigmoid form. The characterization is structural. It identifies the unique admissible modulation that satisfies a maximal-growth constraint. This is a design constraint. Any monotone $[0,1]$-valued modulation that respects the maximal admissible growth must be a sigmoid.

\begin{lemma}[Sigmoid modulation by maximal admissible growth]
\label{lem:sigmoid}
Let $\mathcal{M}$ be the class of $C^1$, strictly monotone functions $m: \mathbb{R} \to [0,1]$ satisfying
\begin{equation}
\begin{aligned}
&\lim_{t \to -\infty} m(t) = 0, \quad \lim_{t \to +\infty} m(t) = 1, \\
&m'(t) > 0 \;\; \text{for all } t \in \mathbb{R}.
\end{aligned}
\label{eq:admissible}
\end{equation}
Within $\mathcal{M}$, consider the maximal admissible growth constraint
\begin{equation}
m'(t) \;\leq\; m(t)\bigl(1 - m(t)\bigr) \quad \text{for all } t \in \mathbb{R}.
\label{eq:growth_constraint}
\end{equation}
Then the sigmoid family
\begin{equation}
m^*(t) = \sigma(\kappa t + \beta), \qquad \kappa \in (0,1],\ \beta \in \mathbb{R},
\label{eq:optmod}
\end{equation}
consists of the functions in $\mathcal{M}$ that attain the constraint~\eqref{eq:growth_constraint} with equality everywhere when $\kappa = 1$. Values $\kappa < 1$ satisfy the constraint strictly. Moreover, for any $m \in \mathcal{M}$ satisfying the strict inequality on a set of positive measure, the deficit
\begin{equation}
\mathcal{D}(m) = \int_{\mathbb{R}} \bigl[m(t)\bigl(1-m(t)\bigr) - m'(t)\bigr]\, dt \;>\; 0,
\label{eq:deficit}
\end{equation}
and $\mathcal{D}(m^*) = 0$ for every $m^*$ in the sigmoid family.
\end{lemma}

\begin{proof}[Proof]
See Appendix~\ref{app:proofs_lemma}.
\end{proof}

Lemma~\ref{lem:sigmoid} settles the choice of modulation function and guarantees that the subsequent evidence modulation preserves routing direction when the temperature scale changes. Given this modulation, we allow the Dirichlet prior itself to depend on $\Delta T$, which goes beyond a purely multiplicative modulation of the base evidence.

Let $e_{\text{base},k}(x)$ be the base evidence from the detection head. The augmented evidence is
\begin{equation}
e_{\text{aug},k}(x) = e_{\text{base},k}(x) \cdot g_k(\Delta T_x) + \pi_k(\Delta T_x),
\label{eq:temp_prior}
\end{equation}
where $g_k(\Delta T) = \mathrm{softplus}(W_k^g \phi(\Delta T))$ and $\pi_k(\Delta T) = \mathrm{softplus}(W_k^\pi \phi(\Delta T))$ are learned temperature-conditioned functions, with $\phi$ a shared two-layer MLP of hidden dimension 32. The modulated evidence is
\begin{equation}
e_{\text{mod},k}(x) = e_{\text{aug},k}(x)\bigl(1 + \gamma\, m(\Delta T_x)\bigr),
\label{eq:evidence}
\end{equation}
where $\gamma > 0$ controls the modulation strength. When $g_k \equiv 1$ and $\pi_k \equiv 0$, Eq.~\eqref{eq:temp_prior} reduces to the base evidence. Note that $\gamma$ (modulation strength) is distinct from the sigmoid slope $\kappa_s$ in Eq.~\eqref{eq:multiscale_mod}.

\begin{proposition}[Exact uncertainty reduction and error propagation]
\label{prop:evidence_modulation}
Let $S_{\text{aug}} = \sum_k (e_{\text{aug},k} + 1)$, $S_{\text{mod}} = \sum_k (e_{\text{mod},k} + 1)$, and define the evidence concentration
\begin{equation}
a = \frac{\bar{e}_{\text{aug}}}{1 + \bar{e}_{\text{aug}}}, \qquad \bar{e}_{\text{aug}} = S_{\text{aug}}/K - 1.
\label{eq:concentration}
\end{equation}
Under the modulation constraints, the uncertainty reduction is given exactly by
\begin{equation}
u_{\text{aug}} - u_{\text{mod}} = \frac{K}{S_{\text{aug}}} \cdot \frac{\gamma\, m(\Delta T)\, a}{1 + \gamma\, m(\Delta T)\, a}.
\label{eq:exact}
\end{equation}
Consequently: (i) the right-hand side is strictly increasing in $\Delta T$ and strictly increasing in $a$ for $\gamma m > 0$; (ii) since $a \in [0,1]$, the uncertainty reduction satisfies the two-sided bound
\begin{equation}
0 \;\le\; u_{\text{aug}} - u_{\text{mod}} \;\le\; \frac{K}{S_{\text{aug}}} \cdot \frac{\gamma\, m(\Delta T)}{1 + \gamma\, m(\Delta T)},
\label{eq:separation}
\end{equation}
with equality on the left iff $a = 0$ and on the right iff $a = 1$; (iii) if $a$ is estimated as $\hat{a}$ with $|\hat{a} - a| \le \epsilon_a$, then
\begin{equation}
\left| (u_{\text{aug}} - u_{\text{mod}}) - \frac{K}{S_{\text{aug}}} \cdot \frac{\gamma m \hat{a}}{1 + \gamma m \hat{a}} \right| \le \frac{K}{S_{\text{aug}}} \cdot \gamma m \cdot \epsilon_a.
\label{eq:errorprop}
\end{equation}
\end{proposition}

\begin{proof}[Proof]
See Appendix~\ref{app:proofs_lemma}.
\end{proof}

Proposition~\ref{prop:evidence_modulation} is the central design equation. It states that the uncertainty reduction is jointly controlled by the physical contrast through $m(\Delta T)$ and by the evidence concentration $a$. Both factors therefore enter the calibration cost. The pairing factor determines whether $a$ is meaningful, and the scale factor determines whether $\Delta T$ is informative.

\begin{corollary}[Affine invariance of temperature evidence]
\label{cor:affine_invariance}
Let $T' = aT + b$ with $a > 0$. Then $\Delta T' = a\,\Delta T$, the sigmoid modulation $m'(\Delta T') = \sigma(\kappa a\,\Delta T + \beta)$ remains admissible in the sense of Lemma~\ref{lem:sigmoid}, the direction of the uncertainty reduction in Proposition~\ref{prop:evidence_modulation} is preserved, and the ordering of $u$ across samples is unchanged. Consequently, a strictly increasing affine reparametrization of the temperature channel preserves routing behavior. Only the physical interpretation of $u$ and the reference scale $T_{\text{ref}}$ in Eq.~\eqref{eq:whigh} are affected.
\end{corollary}

\begin{proof}
See Appendix~\ref{app:proofs_lemma}.
\end{proof}

\paragraph{Shuffled, const-$\Delta T$, and var-phase construction.}
Three attribution probes are constructed from the true $\Delta T$ map.

\emph{Shuffled $\Delta T$.} The shuffled variant is constructed as follows.
(i) The permutation $\pi$ is drawn uniformly from the symmetric group
$\mathfrak{S}_n$ on the target-mask pixels; the symmetric group is not
restricted to derangements, so $\pi$ may contain fixed points.
(ii) For boxes with fewer than two nonzero pixels after resizing, the
original $\Delta T$ values are retained, as no valid permutation exists.
(iii) Shuffling is applied after the shared geometric augmentation, so
the mask geometry is identical between true and shuffled, and only the
within-mask value assignment is permuted.
(iv) At evaluation, boxes are provided by the fixed Stage-1 detector,
matching the oracle-free two-stage protocol; no GT positions are used.
The same $\pi$ is used in training and evaluation for a given image,
seeded by the image id.

\emph{const-$\Delta T$.} The const-$\Delta T$ variant subtracts, per sample
and per dilation scale, the mean of the nonzero $\Delta T$ anchors:
$\Delta T^{(s)} \leftarrow \Delta T^{(s)} - \overline{\Delta T^{(s)}}$
where the mean is taken over anchors with $|\Delta T^{(s)}| > 10^{-6}$.
This preserves the within-image relative temperature structure
(and hence paired correspondence) while removing the absolute
calibrated scale.

\emph{var-phase.} The var-phase variant preserves the per-sample variance and the power-spectrum magnitude and replaces the spatial phase with a uniformly random phase per sample and per dilation scale. The per-sample variance of the true $\Delta T$ map is matched before the phase replacement and restored after it, so the variant isolates spatial phase as the only removed factor while keeping the support domain identical to the true $\Delta T$ map. The random phase is drawn independently per image, seeded by the image id.

\paragraph{Grayscale pseudo-$\Delta T$ construction.}
The pseudo-$\Delta T$ arm uses exactly the same boxes as the true-$\Delta T$ arm (Stage-1 predicted boxes, no ground truth) and the same three dilation scales $r \in \{0.1, 0.2, 0.4\}$. Only the value source differs: where the true arm computes the annular contrast $\Delta T^{(s)}$ on the calibrated temperature matrix $T$, the pseudo arm computes $\Delta G^{(s)}$ on the camera's native 8-bit grayscale rendering $G$, using the same annular geometry as Eq.~\eqref{eq:multiscale}. Both maps then pass through the same per-sample scale normalization
\begin{equation}
\tilde{x} = \frac{x}{\mathrm{std}(x_{\text{nonzero}})} \cdot \sigma_0,
\label{eq:scale_norm}
\end{equation}
where $x_{\text{nonzero}}$ denotes the anchors with $|x| > 10^{-6}$ across all three dilation scales, and $\sigma_0 = 0.3804$ is an empirically fixed target scale shared by every structured arm (true, pseudo, random, shuffled, var-phase). After normalization, the nonzero standard deviation of every arm is exactly $\sigma_0$, so true and pseudo enter TEL at the same second-moment scale. The two arms therefore differ only in the information source of the injected contrast, not in the geometry, support, or scale of the modulation. The parameter and FLOP counts are identical because both arms share the same TEL head. Table~\ref{tab:deltaTb} therefore marks them as matched. The normalized $\Delta G$ and $\Delta T$ maps agree in second-moment scale but not in mean or higher moments. The true-vs-pseudo contrast therefore measures the information content of the calibrated channel at matched scale, not at matched distribution. The value $\sigma_0 = 0.3804$ is fixed across arms and datasets; it sets the input scale of the modulation gate and does not affect the true-vs-pseudo contrast, which compares two arms at the same $\sigma_0$.

\subsection{Information-Theoretic Layer: KL Separability of the Temperature Channel}
\label{sec:separability}
\begin{figure*}[!t]
    \centering
    \begin{minipage}[c]{0.49\textwidth}
        \centering
        \begin{subfigure}[b]{\textwidth}
            \centering
            \includegraphics[width=\textwidth]{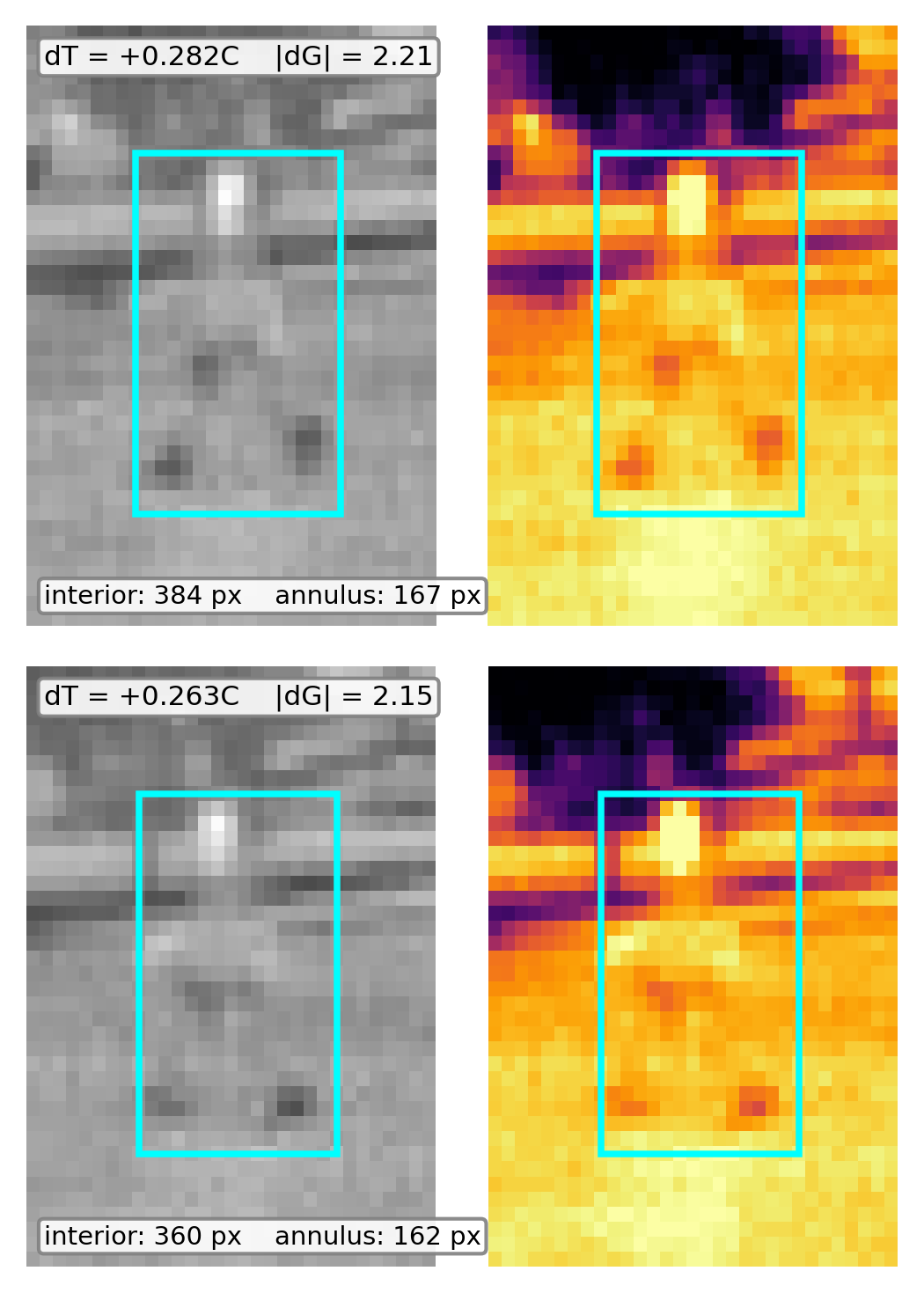}
            \caption{Nighttime patches.}
            \label{fig:deltaT_physics_a}
        \end{subfigure}
    \end{minipage}
    \hfill
    \begin{minipage}[c]{0.50\textwidth}
        \centering
        \begin{subfigure}[b]{\textwidth}
            \centering
            \includegraphics[width=\textwidth]{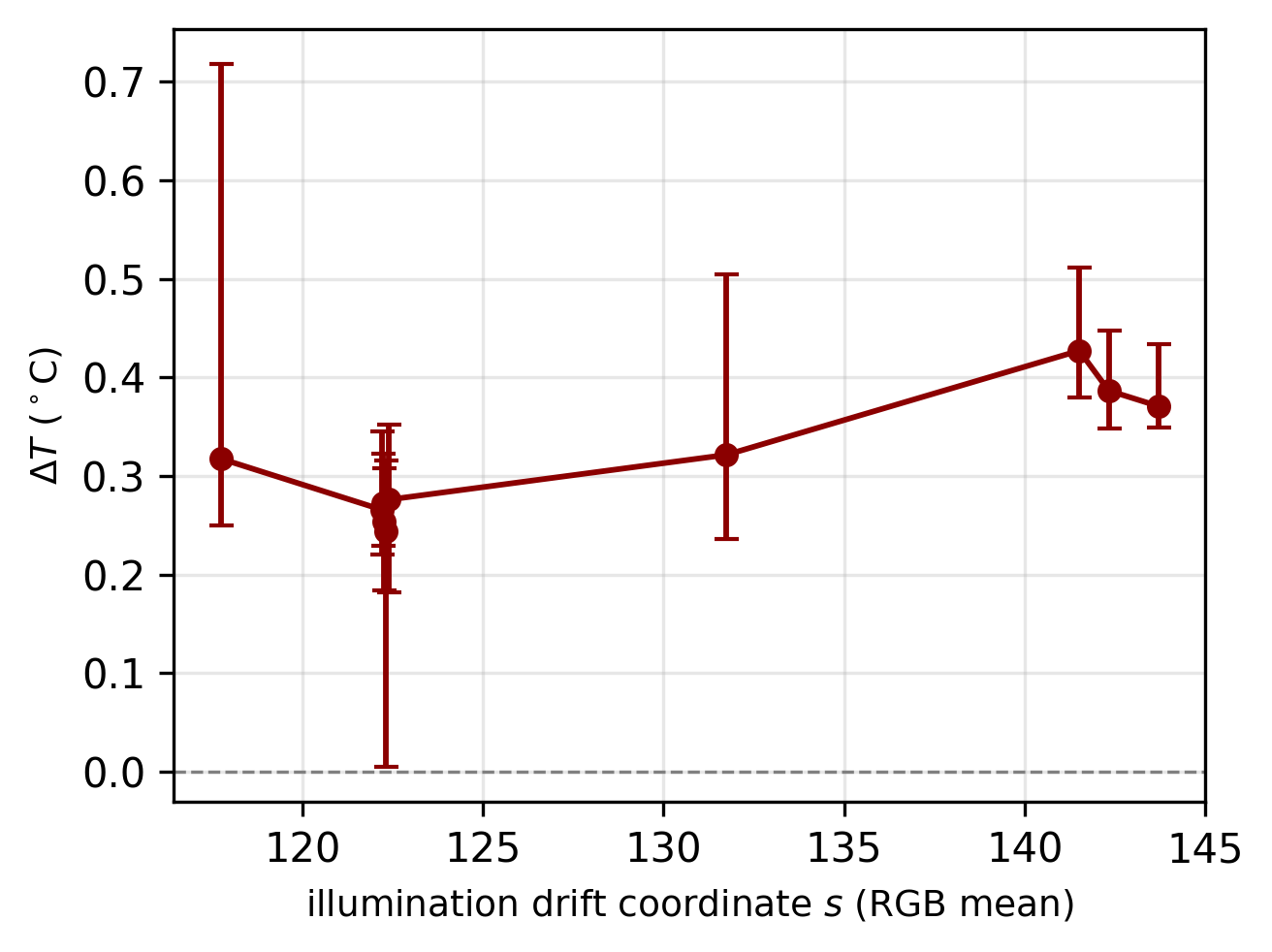}
            \caption{Illumination drift.}
            \label{fig:deltaT_physics_b}
        \end{subfigure}
        \\[4pt]
        \begin{subfigure}[b]{\textwidth}
            \centering
            \includegraphics[width=\textwidth]{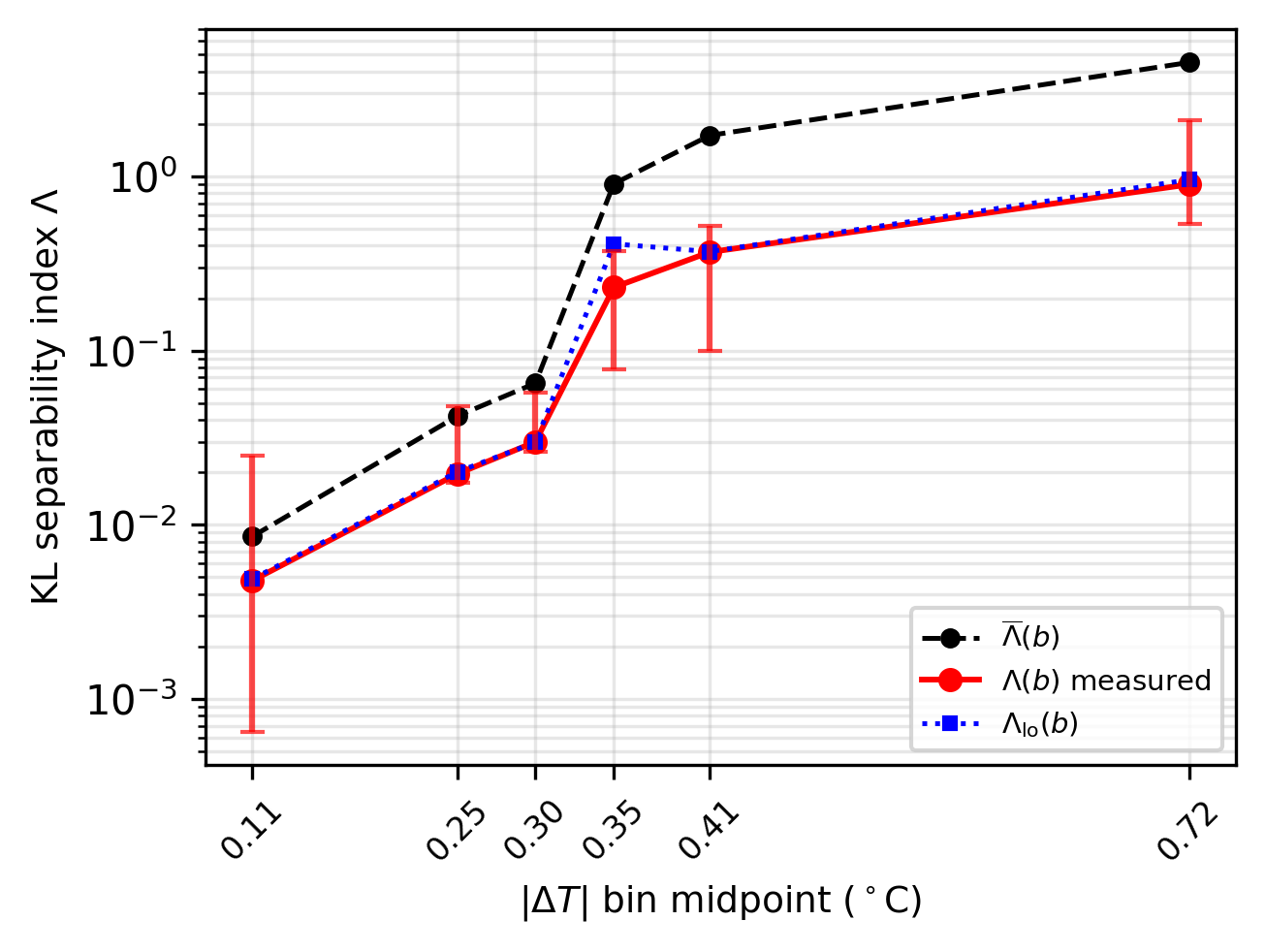}
            \caption{KL separability sandwich.}
            \label{fig:deltaT_physics_c}
        \end{subfigure}
    \end{minipage}
    \caption{Physical evidence of the temperature matrix.
    (a)~Two nighttime patches in which the grayscale contrast is low ($|\Delta G| \approx 2.2$) while the calibrated $\Delta T$ remains measurable ($\approx 0.28^\circ$C). The inset text reports the exact number of interior and annular background pixels.
    (b)~Conditional distribution $P(\Delta T \mid y{=}\text{person})$ as a function of the illumination drift coordinate $s$, stratified into quintiles. $P(y\mid s)$ is stationary, while $P(\Delta T \mid y,s)$ drifts with $s$. Data points mark median $\pm$ IQR.
    (c)~Empirical validation of the KL separability sandwich in Eq.~\eqref{eq:kl_sandwich}. The red line is the measured median $\Lambda(b)$ with interquartile range (IQR) error bars, obtained by using the thermal camera's native 8-bit grayscale rendering (strictly pixel-aligned with the 14-bit TIFF) to compute the true local grayscale ratio $x$. The black dashed line is the theoretical upper bound $\overline{\Lambda}(b)$. The empirical median grayscale-to-temperature noise ratio is $x \approx 1.10$. The empirical median strictly falls within the theoretical bounds, confirming that the calibrated temperature channel $\Delta T$ provides a reliable source of local separability evidence. The empirical median temperature noise is $\sigma_{med}=0.39^\circ$C (IQR: $[0.22, 0.72]^\circ$C).}
    \label{fig:deltaT_physics}
\end{figure*}
Proposition~\ref{prop:evidence_modulation} proves that $\Delta T$ reduces $u$, but it does not explain why $\Delta T$ carries separable information. This gap motivates the information-theoretic layer. To address it, we work directly with the continuous temperature channel and adopt a KL separability index. Under the local Gaussian shift model standard for calibrated radiometric sensors, the index is closed-form and unbounded in Signal-to-Noise Ratio (SNR). We use this model as a diagnostic. Fig.~\ref{fig:deltaT_physics} provides the empirical basis: panel~(a) shows the physical motivation, panel~(b) supports the local Gaussian shift model of Definition~\ref{def:gaussian_shift}, and panel~(c) tests the bound.

\begin{definition}[Local Gaussian shift model]
\label{def:gaussian_shift}
Fix a candidate box $b$ with person indicator $y_b \in \{0,1\}$ and prior $\pi_b = \Pr(y_b = 1)$. The annular contrast $Z_b = \Delta T_b$ satisfies
\begin{equation}
Z_b \mid y_b = 0 \sim \mathcal{N}(0, \sigma^2), \qquad
Z_b \mid y_b = 1 \sim \mathcal{N}(\mu, \sigma^2),
\label{eq:gaussian_shift}
\end{equation}
with $\sigma^2 = \sigma_T^2 + s^{-2}\sigma_G^2$, where $\sigma_T^2$ is the sensor temperature noise, $s>0$ is the local slope of the grayscale rendering $G = \phi(T)$ on the relevant interval, and $s^{-1}\sigma_G$ is the grayscale readout noise. Denote $\SNR_b^{\Delta T} = \mu^2/\sigma_T^2$ and $\SNR_b^{G} = \mu^2/\sigma^2$.
\end{definition}

\begin{definition}[KL separability index]
\label{def:kl_index}
The KL separability index of the temperature matrix at box $b$ is
\begin{equation}
\Lambda(b) = \KL\!\left(p(Z_b \mid y_b = 1) \,\|\, p(Z_b \mid y_b = 0)\right).
\label{eq:kl_index}
\end{equation}
\end{definition}

\begin{proposition}[KL separability sandwich]
\label{prop:separability_sandwich}
Under Definitions~\ref{def:gaussian_shift}--\ref{def:kl_index}, the following hold.

\emph{(i) Closed form and non-saturating bounds.}
\begin{equation}
\Lambda(b) = \tfrac{1}{2}\SNR_b^{G}
= \tfrac{1}{2}\cdot\frac{\mu^2}{\sigma_T^2 + s^{-2}\sigma_G^2},
\label{eq:kl_closed}
\end{equation}
and consequently
\begin{equation}
\underbrace{\tfrac{1}{2}\SNR_b^{\Delta T}\cdot\frac{1}{1+x}}_{\underline{\Lambda}(b)}
\;\le\; \Lambda(b) \;\le\;
\underbrace{\tfrac{1}{2}\SNR_b^{\Delta T}}_{\overline{\Lambda}(b)},
\label{eq:kl_sandwich}
\end{equation}
where \(x = s^{-2}\sigma_G^2/\sigma_T^2 \in [0,\infty)\). The lower bound is non-negative for all \(x \ge 0\). When the grayscale rendering collapses (\(x \to 0\)), the two bounds converge and the sandwich is tight. When grayscale information is abundant (\(x \to \infty\)), the upper bound stays fixed while the lower bound decays as \(1/(1+x)\), so the sandwich becomes loose; equivalently, \(\Lambda(b)=\tfrac12\SNR_b^{\Delta T}/(1+x)=\tfrac12\SNR_b^G\), and the separability is then governed by the grayscale channel. The ratio \(\overline{\Lambda}/\underline{\Lambda}=1+x\) diverges as \(x\to\infty\).
\emph{(ii) Exponential Bayes-error decay in the idealized model.}
Let \(P_e^{G,\Delta T}(b)\) be the Bayes error on \(b\) in the local Gaussian shift model using both \(G_X|_b\) and \(\Delta T_b\). Then
\begin{equation}
P_e^{G,\Delta T}(b) \;\le\; \tfrac{1}{2} e^{-\Lambda(b)}.
\label{eq:bayes_decay}
\end{equation}
\emph{(iii) Coupling to cognitive uncertainty.}
Let \(S_0 = \sum_k (e_{\text{base},k} + 1)\) and \(u_0 = K/S_0\). The uncertainty reduction induced by temperature-evidential modulation satisfies
\begin{equation}
u_0 - u_{\Delta T}
\;=\; \frac{K}{S_0}\bigl(1 - e^{-\Lambda(b)}\bigr) + O\!\left(\SNR_b^{-1}\right).
\label{eq:u_kl_coupling}
\end{equation}
\end{proposition}

\begin{proof}[Proof]
See Appendix~\ref{app:proofs_separability}.
\end{proof}

The exponential decay bound in Proposition~\ref{prop:separability_sandwich}(ii) is used as a diagnostic under the local Gaussian shift model.
\begin{remark}[KL index as a local diagnostic]
\label{rem:kl_diagnostic}
The KL separability index is preferred over a binary mutual-information formulation because the latter saturates at one bit under the high signal-to-noise ratios typical of calibrated thermal imagery, whereas $\Lambda(b)$ grows linearly in $\SNR_b$ and remains closed-form. The empirical counterpart is illustrated in
Fig.~\ref{fig:deltaT_physics}(c) and Table~\ref{tab:supp_s3}.
\end{remark}

\subsection{Decision Layer: Unified Routing via Evidential Scoring} 
\label{sec:routing}

The physical and information-theoretic layers establish that the paired thermal channel carries separable information and reduces cognitive uncertainty. The decision layer uses this information for annotation, pseudo-labeling, and discard through one routing rule. We evaluate routing as a design choice, and Section~\ref{sec:discussion} includes an exploratory probe.

Fix $\tau_{\text{low}} < \tau_{\text{high}}$ and route by
\begin{equation}
\rho(x) = \begin{cases}
\text{annotate} & u(x) > \tau_{\text{high}}, \\
\text{pseudo} & \tau_{\text{low}} \le u(x) \le \tau_{\text{high}}, \\
\text{discard} & u(x) < \tau_{\text{low}},
\end{cases}
\label{eq:routing}
\end{equation}
where $u(x)=K/S(x)$ is the cognitive uncertainty proxy. Proposition~\ref{prop:routing_bound}(i) gives the thresholds an interpretation, where $\tau_{\text{high}}$ is the smallest uncertainty for which annotation is worth the cost, and $\tau_{\text{low}}$ is the largest uncertainty for which a pseudo-label is safe to accept. Since $u$ is a scalar, a sample lies in exactly one routing branch, so its annotation, pseudo-labeling, and discard decisions are consistent across rounds. Fig.~\ref{fig:lifecycle}(c) shows where the scalar enters the closed loop.

A fixed-threshold router is sensitive to the absolute scale of $u$. We therefore also consider the quantile router
\begin{equation}
\rho_Q(x) = \begin{cases}
\text{annotate} & u(x) > Q_{1-\alpha}(u), \\
\text{pseudo} & Q_{\beta}(u) \le u(x) \le Q_{1-\alpha}(u), \\
\text{discard} & u(x) < Q_{\beta}(u),
\end{cases}
\label{eq:quantile_routing}
\end{equation}
whose invariance follows from Corollary~\ref{cor:quantile_invariance}. The empirical three-state structure of $u$ under this router is analyzed in Section~\ref{sec:routing-scalar-structure}. Routing is based on $u$, which is strictly monotone in the total evidence $S$. The predictive entropy $H(\hat{p})$ is used as a diagnostic and does not participate in routing. Remark~\ref{rem:q1_validation} quantifies the empirical relation between $u$ and $H$. The routing layer separates two quantities with distinct roles. See Section~\ref{sec:routing-scalar-structure} for the empirical characterization.

\begin{proposition}[Unified routing: acquisition proxy, pseudo-label error, and generalization]
\label{prop:routing_bound}
Under a Dirichlet predictive distribution with total evidence \(S(x)\):
\begin{enumerate}[label=(\roman*), leftmargin=1.5em]
\item \emph{Information-gain proxy.}
Define the routing acquisition score
\begin{equation}
\Delta_{\mathrm{acq}}(x)
= \frac{K}{S(x)} \cdot \Psi(x)
= u(x) \cdot \Psi(x),
\label{eq:infogain2}
\end{equation}
with
$
\Psi(x)
= \mathbb{E}_{p \sim \mathrm{Dir}(\alpha)}[H(p)] > 0,
$
where \(\alpha_k = e_k + 1\) are the Dirichlet parameters and \(H\)
is the Shannon entropy. We treat this score as a monotone acquisition proxy.
In the large-evidence limit the exact information gain satisfies
\begin{equation}
\begin{aligned}
\lefteqn{\mathrm{IG}(x) = \mathbb{E}[H(p)] - \mathbb{E}_{y}\mathbb{E}[H(p\mid y)] }\\
&= \frac{K-1}{2S(x)} + O(S(x)^{-2})
= \frac{K-1}{2K}u(x) + O(S(x)^{-2}),
\end{aligned}
\label{eq:exact_ig_asym}
\end{equation}
so ranking by \(u(x)\) is consistent with ranking by the exact
information gain in that limit.
\item \emph{Pseudo-label error under correlation.} Suppose $f^{(t)}$ and $f^{(t-1)}$ each have error probability $\varepsilon$ on the medium-uncertainty band, and let $\rho$ denote the Pearson correlation coefficient between their indicator errors. Under the consensus rule
\begin{equation}
\begin{aligned}
\mathrm{accept~}\hat{y}(x) \iff\;&
\mathrm{IoU}\bigl(b^{(t)}(x), b^{(t-1)}(x)\bigr) > 0.8 \\
&\wedge\; \hat{y}^{(t)}(x) = \hat{y}^{(t-1)}(x),
\end{aligned}
\label{eq:consensus}
\end{equation}
the accepted pseudo-label error satisfies
\begin{equation}
\Pr(\hat{y} \neq y \mid \text{accept})
\;\le\;
\frac{\varepsilon^{2} + \rho\,\varepsilon(1-\varepsilon)}
     {\varepsilon^{2} + (1-\varepsilon)^{2} + 2\rho\,\varepsilon(1-\varepsilon)}.
\label{eq:quadratic}
\end{equation}
At $\rho = 0$ this equals $\varepsilon^{2}/[\varepsilon^{2} + (1-\varepsilon)^{2}]$; at $\rho = 1$ it equals $\varepsilon$.
\item \emph{Non-i.i.d.\ active-routing generalization.} Assume the base detector has bounded loss in $[0,1]$, and the routing propensity $e(x) = \Pr(\rho(x) = \text{annotate} \mid x)$ satisfies $e(x) \ge e_{\min} > 0$. Define the inverse-propensity-weighted (IPW) risk estimator
\begin{equation}
\hat{L}_{\mathrm{IPW}} = \frac{1}{n}\sum_{i=1}^{n}
\frac{\mathbb{1}[\rho(x_i) = \text{annotate}]}{e(x_i)}\,\ell(f(x_i), y_i).
\label{eq:ipw}
\end{equation}
Then with probability at least $1 - \delta$,
\begin{equation}
\begin{aligned}
L_{\text{test}} \;\le\;& \hat{L}_{\mathrm{IPW}} + \sqrt{\frac{\log(2/\delta)}{2 n\, e_{\min}^{2}}} \\
&+ \frac{n_P}{n}\, \varepsilon_{\text{pseudo}} + \frac{n_D}{n}\, \varepsilon_{\text{discard}}(\tau_{\text{low}}),
\end{aligned}
\label{eq:router_gen}
\end{equation}
where $\varepsilon_{\text{pseudo}}$ is the pseudo-label error bound above. Under $\phi$-mixing with $\phi(m) \le c\rho^{m}$, the Hoeffding term is replaced by
\begin{equation}
\sqrt{\frac{\log(2/\delta)}{2 n\, e_{\min}^{2}}} + 2\sum_{m=1}^{n}\phi(m)
= O\!\left(\sqrt{\frac{\log(2/\delta)}{n\, e_{\min}^{2}}}\right) + O(1).
\label{eq:mixing}
\end{equation}
\item \emph{Online-routing regret.} If the routing loss $\ell_t(\rho_t)$ is convex and $G$-Lipschitz, online gradient descent on the routing parameters satisfies
\begin{equation}
\begin{aligned}
&\sum_{t=1}^{T}\ell_t(\rho_t) - \min_{\rho}\sum_{t=1}^{T}\ell_t(\rho) \\
&\quad\le\; \frac{\|\rho_0 - \rho^*\|^{2}}{2\eta} + \frac{\eta G^{2} T}{2} = O(\sqrt{T}).
\end{aligned}
\label{eq:regret}
\end{equation}
\end{enumerate}
\end{proposition}

\begin{proof}[Proof]
See Appendix~\ref{app:proofs_routing}.
\end{proof}

\begin{proposition}[Finite-evidence degeneracy of the acquisition product]
\label{prop:finite-S-degeneracy}
Let $u=K/S$, $\Psi=\mathbb{E}_{p\sim\mathrm{Dir}(\alpha)}[H(p)]$, and $\Delta_{\mathrm{acq}}=u\cdot\Psi$. Write $\Psi(x)=\bar{\Psi}+\delta(x)$ with $\mathbb{E}[\delta]=0$. Suppose the entropy variance contracts relative to the routing variance,
$
\eta_S = \frac{\mathbb{E}[u^{2}]\,\mathrm{Var}(\Psi)}
              {\bar{\Psi}^{2}\,\mathrm{Var}(u)}
       \to 0,
$
and the entropy range satisfies $\mathrm{CV}(\Psi)=O(\eta_S^{1/2})$. Then
$
\Delta_{\mathrm{acq}}(x) = \bar{\Psi}\,u(x) + u(x)\,\delta(x),
$
and the residual term is controlled by
\[
\frac{|\Delta_{\mathrm{acq}}(x)-\bar{\Psi}\,u(x)|}{|\bar{\Psi}\,u(x)|}
= \frac{|\delta(x)|}{|\bar{\Psi}|}.
\]
Consequently, the ranking induced by $\Delta_{\mathrm{acq}}$ coincides with that of $u$ up to $O(\eta_S^{1/2})$ perturbations, and the coefficient of determination of the linear fit $\Delta_{\mathrm{acq}}=a u+b$ satisfies $R^2\geq 1-\eta_S$.
\end{proposition}
\begin{corollary}[Finite-evidence degeneration of the acquisition product]
\label{cor:acq_degeneration}
Under the decomposition of Proposition~\ref{prop:finite-S-degeneracy}, the following hold.

\emph{(i) Pointwise relative error.} For every $x$,
\[
\frac{|\Delta_{\mathrm{acq}}(x) - \bar{\Psi}\,u(x)|}{|\bar{\Psi}\,u(x)|}
= \frac{|\delta(x)|}{|\bar{\Psi}|}.
\]
Hence $\Delta_{\mathrm{acq}}$ is a relative perturbation of $\bar{\Psi}\,u$ of order $|\delta(x)|/|\bar{\Psi}|$ at each $x$.

\emph{(ii) Rank preservation.} Assume $|\delta(x)|\leq\varepsilon|\bar{\Psi}|$ for all $x$, with $\varepsilon\in(0,1)$. Then for any pair $(x,y)$ with
$
\frac{u(x)}{u(y)} > \frac{1+\varepsilon}{1-\varepsilon},
$
the ordering is preserved:
$$
\Delta_{\mathrm{acq}}(x) > \Delta_{\mathrm{acq}}(y)
\iff u(x) > u(y).
$$
Rank flips can occur only between pairs whose relative $u$-gap is at most $2\varepsilon/(1-\varepsilon)$.

\emph{(iii) Coefficient of determination.} If $\delta$ is uncorrelated with $u$, the linear regression of $\Delta_{\mathrm{acq}}$ on $u$ attains
$R^{2} = \frac{1}{1+\eta},$
and $\eta = \frac{\mathbb{E}[u^{2}]\,\mathrm{Var}(\Psi)}
             {\bar{\Psi}^{2}\,\mathrm{Var}(u)}.$
In particular, $R^{2}\to 1$ as $\eta\to 0$, and $R^{2}\geq 1-\eta$.

\emph{(iv) Empirical values on FLIR ADAS.} Over $n=5.4\times 10^{5}$ pooled RoIs and ten seeds,
$\frac{\mathrm{Var}(\Psi)}{\mathrm{Var}(u)} = 8.8\times 10^{-3},$
$\frac{\mathrm{std}(\Psi)}{\mathrm{std}(u)} = 9.4\times 10^{-2},$
and
$
\eta = 6.5\times 10^{-3}.
$
The predicted value is $R^{2}=0.9936$; the measured value is $R^{2}=0.9958$ with slope $\bar{\Psi}=0.4982$ and residual standard deviation $0.007054$. The acquisition product therefore degenerates to $u$ in this regime.
\end{corollary}

\begin{proof}
See Appendix~\ref{app:finite-S-degeneracy}.
\end{proof}

\begin{remark}[Empirical validation of the pseudo-label error bound, Eq.~\eqref{eq:quadratic}]
\label{rem:q2b_validation}
We validate Eq.~\eqref{eq:quadratic} on the medium-uncertainty band induced by each of four signals: the predictive entropy \(H(\hat{p})\), the class confidence, the top-two margin, and \(u=K/S\). For each signal and each pair of seeds \((s_i,s_j)\), \(i<j\), we estimate on the accepted subset (i) the single-model error rate \(\varepsilon\), (ii) the Pearson correlation \(\rho\) between the two models' error indicators, and (iii) the empirical accepted error \(\Pr(\hat{y}\neq y\mid\text{accept})\), with correctness determined by ground truth and a class mismatch counted as an error.

\begin{table}[t]
\centering
\caption{Empirical validation of Eq.~\eqref{eq:quadratic} over all $45$ seed pairs. Margin and conf are identical to each other; $u$ is numerically
indistinguishable from them because $K=2$. Coverage is the fraction of seed pairs for which the bound exceeds the accepted error.}
\label{tab:oracle_bound}
\small
\setlength{\tabcolsep}{3pt}
\resizebox{\columnwidth}{!}{%
\begin{tabular}{lccccc}
\toprule
Signal & $\varepsilon$ & $\rho$ & $\text{err}|\text{acc}$ & $\text{ub}$ & Coverage \\
\midrule
\multicolumn{6}{l}{\textit{Mean $\pm$ std across the $45$ seed pairs}} \\
$u$ / margin / conf & $0.086 \pm 0.021$ & $0.712 \pm 0.072$ & $0.0364 \pm 0.014$ & $0.0660 \pm 0.016$ & $45/45$ \\
$H$ & $0.116 \pm 0.021$ & $0.520 \pm 0.084$ & $0.0281 \pm 0.010$ & $0.0736 \pm 0.014$ & $45/45$ \\
\midrule
\multicolumn{6}{l}{\textit{$95\%$ bootstrap CI (2000 resamples, pair as unit)}} \\
$u$ / margin / conf & $[0.080, 0.092]$ & $[0.690, 0.734]$ & $[0.032, 0.041]$ & $[0.061, 0.071]$ & --- \\
$H$ & $[0.110, 0.122]$ & $[0.495, 0.545]$ & $[0.025, 0.031]$ & $[0.070, 0.078]$ & --- \\
\bottomrule
\end{tabular}%
}
\vspace{2pt}
\end{table}
As Table~\ref{tab:oracle_bound} shows, for $u$, margin, and conf, the bound exceeds the observed accepted error on all $45$ seed pairs, with a mean point-estimate gap of $+0.030$. For $H$, the bound also exceeds the observed accepted error on all $45$ seed pairs, with a mean gap of $+0.046$. Thus Eq.~\eqref{eq:quadratic} provides an empirically valid upper bound for both signal families. The three signals $u$, margin, and conf behave similarly across seed pairs. Their redundancy under $K=2$ is quantified in Table~\ref{tab:q1}.
\end{remark}

\begin{table}[t]
\centering
\caption{Signal correlations on the FLIR ADAS prediction protocol, pooled over ten seeds. Values are Pearson $\rho$ and Spearman $\rho$ with $95\%$ bootstrap confidence intervals on the pooled RoIs.}
\label{tab:q1}
\small
\setlength{\tabcolsep}{3pt}
\resizebox{\columnwidth}{!}{%
\begin{tabular}{lcccc}
\toprule
Pair & Pearson $\rho$ & 95\% CI & Spearman $\rho$ & 95\% CI \\
\midrule
\multicolumn{5}{l}{\textit{Signal-to-signal correlations}} \\
$u$, $H$ & $-0.0765$ & $[-0.0782,-0.0748]$ & $-0.2662$ & $[-0.2687,-0.2639]$ \\
$u$, margin & $-0.9255$ & $[-0.9263,-0.9247]$ & $-0.3079$ & $[-0.3109,-0.3047]$ \\
$u$, conf & $-0.9255$ & $[-0.9263,-0.9247]$ & $-0.3079$ & $[-0.3110,-0.3050]$ \\
$H$, margin & $-0.2590$ & $[-0.2608,-0.2573]$ & $-0.6202$ & $[-0.6224,-0.6180]$ \\
$H$, conf & $-0.2590$ & $[-0.2608,-0.2573]$ & $-0.6202$ & $[-0.6225,-0.6180]$ \\
margin, conf & $+1.0000$ & $[+1.0000,+1.0000]$ & $+1.0000$ & $[+1.0000,+1.0000]$ \\
\midrule
\multicolumn{5}{l}{\textit{Correlations with $\Delta = u\cdot\mathbb{E}[H(p)]$}} \\
$\Delta$, $H$ & $-0.0151$ & $[-0.0167,-0.0135]$ & $+0.4142$ & $[+0.4112,+0.4173]$ \\
$\Delta$, $u$ & $+0.9979$ & $[+0.9979,+0.9979]$ & $+0.5444$ & $[+0.5418,+0.5470]$ \\
$\Delta$, margin & $-0.9455$ & $[-0.9462,-0.9450]$ & $-0.8818$ & $[-0.8832,-0.8803]$ \\
$\Delta$, conf & $-0.9455$ & $[-0.9461,-0.9449]$ & $-0.8818$ & $[-0.8831,-0.8804]$ \\
\bottomrule
\end{tabular}%
}
\end{table}

\begin{corollary}[Model-conditional monotone discard-risk bound]
\label{cor:discard_monotone}
Under the Dirichlet evidence model, with total evidence $S$, cognitive uncertainty proxy $u = K/S$, and $K$ classes,
\begin{equation}
\begin{aligned}
\varepsilon_{\text{discard}}(\tau_{\text{low}})
&= \mathbb{E}\!\left[\mathbb{1}[\hat{y}(x) \neq y] \,\big|\, u(x) < \tau_{\text{low}}\right] \\
&\le\; \frac{K-1}{K/\tau_{\text{low}} + 1},
\end{aligned}
\label{eq:discard_bound}
\end{equation}
and the right-hand side is strictly increasing in $\tau_{\text{low}}$.
\end{corollary}

\begin{proof}
See Appendix~\ref{app:proofs_routing}.
\end{proof}

\begin{remark}[Empirical validation of the discard-risk bound]
\label{rem:q2c_validation}
We validate Corollary~\ref{cor:discard_monotone} on an oracle-based
image-level protocol, averaged over ten seeds. For each ground-truth
person box, the nearest detection's cognitive uncertainty $u$ serves as
the routing score, and correctness is determined by IoU $> 0.5$ against
the same ground truth. This image-level protocol avoids the
background-false-positive inflation that dominates a per-RoI evaluation,
where over $85\%$ of detections are background.

Table~\ref{tab:q2c} summarizes the mean measured discard error
$\Pr(\hat{y}\neq y \mid u < \tau_{\text{low}})$ over the ten seeds. The
measured error stays below the model-conditional bound
$(K-1)/(K/\tau_{\text{low}}+1)$ at every $\tau_{\text{low}}$, and the
bound is strictly increasing in $\tau_{\text{low}}$, matching
Corollary~\ref{cor:discard_monotone}.
\end{remark}

\begin{table}[t]
\centering
\caption{Empirical validation of Corollary~\ref{cor:discard_monotone} on an oracle-based image-level protocol, averaged over ten seeds. Thresholds below $0.40$ are omitted because the discard branch contains too few samples for a reliable estimate.}
\label{tab:q2c}
\small
\resizebox{\columnwidth}{!}{%
\begin{tabular}{lccc}
\toprule
$\tau_{\text{low}}$ & measured (mean $\pm$ std) & $[\min,\max]$ & bound \\
\midrule
$0.40$ & $0.0802 \pm 0.0704$ & $[0.0126, 0.2231]$ & $0.1667$ \\
$0.45$ & $0.0308 \pm 0.0158$ & $[0.0129, 0.0592]$ & $0.1837$ \\
$0.50$ & $0.0237 \pm 0.0058$ & $[0.0162, 0.0378]$ & $0.2000$ \\
$0.55$ & $0.0332 \pm 0.0051$ & $[0.0254, 0.0454]$ & $0.2157$ \\
$0.60$ & $0.0470 \pm 0.0027$ & $[0.0423, 0.0524]$ & $0.2308$ \\
\bottomrule
\end{tabular}%
}
\end{table}

Proposition~\ref{prop:routing_bound} and Corollary~\ref{cor:discard_monotone} demonstrate that the single scalar $u$ carries the information needed to consistently drive annotation, pseudo-labeling, and discard. Proposition~\ref{prop:order_consistency} states the fixed-shape idealization under which $u$ is the unique ranking signal. The subsequent remark records the empirical relation between $u$ and the predictive entropy in the actual TEL evidence distribution.

\begin{proposition}[Fixed-shape idealization of the routing scalar]
\label{prop:order_consistency}
Under Dirichlet evidence with total strength \(S=\sum_{k=1}^K \alpha_k\) and fixed concentration shape \(p_k\), i.e., \(\alpha_k = S p_k\) with \(\sum_{k=1}^K p_k = 1\), the predictive distribution is \(\hat{p}_k = p_k\), independent of \(S\). The predictive entropy \(H(\hat{p})\) and the top-two margin \(m(\hat{p})=\hat{p}_{(1)}-\hat{p}_{(2)}\) are constant in \(S\), while the cognitive uncertainty \(u=K/S\) is strictly decreasing in \(S\). Consequently, \(u\) is the only quantity that varies with \(S\) and thus provides the unique ranking information over samples.
\end{proposition}

\begin{proof}
See Appendix \ref{app:proofs_sufficiency}.
\end{proof}

\begin{remark}[Empirical relation between \(u\) and \(H\) in the actual TEL evidence distribution]
\label{rem:q1_validation}
 In the actual TEL evidence distribution, the shape is not fixed, and \(u\) and the predictive entropy \(H(\hat{p})\) are not comonotone. The reason is that in the two-class case (person vs.\ background), \(S=e_0+e_1+2\). A concentrated evidence profile (one of \(e_0,e_1\) large, the other small) gives moderate \(S\) and small \(H\). A flat evidence profile (both \(e_0,e_1\) large) gives large \(S\) and \(H\) near \(\log 2\). Larger \(S\) therefore does not imply greater certainty in TEL. It often means the model is hesitating between two classes.

We validate this empirically on the FLIR ADAS prediction protocol, pooling the RoIs of ten seeds. Table~\ref{tab:q1} tabulates the pooled Pearson and Spearman correlations among the four candidate signals, together with their correlations with the information-gain product $\Delta=u\cdot\mathbb{E}[H(p)]$. The 95\% confidence intervals are obtained by bootstrapping the pooled RoIs with $1000$ resamples.

The empirical pattern confirms that the fixed-shape idealization of Proposition~\ref{prop:order_consistency} does not transfer to TEL's actual evidence distribution: $u$ and $H(\hat{p})$ are weakly anti-correlated in value ($\rho_{\text{Pearson}}=-0.077$) and moderately anti-correlated in rank ($\rho_{\text{Spearman}}=-0.266$). The top-two margin and the class confidence are perfectly correlated, reflecting a $K=2$ redundancy. The information-gain product $\Delta=u\cdot\mathbb{E}[H(p)]$ is nearly collinear with $u$ ($\rho_{\text{Pearson}}=+0.998$) and essentially uncorrelated with $H$ ($\rho_{\text{Pearson}}=-0.015$); it is a rescaling of $u$ instead of a compromise between $u$ and $H$. Routing therefore reduces to $u$ alone, and the entropy terms carry no independent ordering information.
\end{remark}

\begin{corollary}[Quantile router invariance]
\label{cor:quantile_invariance}
Let $T' = aT + b$ with $a > 0$. Then the quantile router~\eqref{eq:quantile_routing} satisfies
\begin{equation}
\{x : u'(x) > Q_{1-\alpha}(u')\} = \{x : u(x) > Q_{1-\alpha}(u)\},
\label{eq:quantile_invariance}
\end{equation}
and similarly for the pseudo and discard branches. Consequently the routing partition is invariant to strictly increasing affine reparametrizations of the temperature channel.
\end{corollary}

\begin{proof}
See Appendix~\ref{app:proofs_sufficiency}.
\end{proof}

\begin{remark}[Empirical validation of quantile-router invariance]
\label{rem:q3_validation}
We verify Corollary~\ref{cor:quantile_invariance} on the trained detector by applying four strictly increasing affine reparametrizations of the temperature channel, $(a,b)\in\{(1,0),(0.5,0),(1,10),(2,-5)\}$, and comparing the resulting routing partitions to the identity case $(a,b)=(1,0)$. For each reparametrization, the routing consistency $\kappa_{\text{route}}\in[0,1]$ measures the fraction of samples that retain their branch, and $(q_{\text{lo}},q_{\text{hi}})$ are the induced quantile thresholds on $u$.

\begin{table}[t]
\centering
\caption{Empirical validation of Corollary~\ref{cor:quantile_invariance}. Four strictly increasing affine reparametrizations of the temperature channel, $T'=aT+b$, are applied; the routing partition is unchanged in all four cases.}
\label{tab:q3}
\small
\resizebox{\columnwidth}{!}{%
\begin{tabular}{cccc}
\toprule
$a$ & $b$ & consistency $\kappa_{\text{route}}$ & $(q_{\text{lo}},q_{\text{hi}})$ \\
\midrule
$1.0$ & $0.0$ & $1.0000$ & $(0.4520,0.9883)$ \\
$0.5$ & $0.0$ & $1.0000$ & $(0.4520,0.9883)$ \\
$1.0$ & $10.0$ & $1.0000$ & $(0.4520,0.9883)$ \\
$2.0$ & $-5.0$ & $1.0000$ & $(0.4520,0.9883)$ \\
\bottomrule
\end{tabular}%
}
\end{table}

The routing consistency under each reparametrization is recorded in Table~\ref{tab:q3}. All four reparametrizations give $\kappa_{\text{route}}=1.0000$ and identical quantile thresholds to three decimal places, confirming Corollary~\ref{cor:quantile_invariance} empirically.
\end{remark}

\subsection{Scale Structure and Two-Ended Separation of the Routing Scalar}
\label{sec:routing-scalar-structure}

The routing layer adopts $u=K/S$ as a scalar router and retains $\Psi=\mathbb{E}[H(p)]$ as a diagnostic. We characterize the empirical geometry of this layer through scale invariance, variance contraction, partial-correlation control, and two-ended separation. Together, these four properties indicate that the router does not require a monotone uncertainty ranking.

The quantile router is invariant under any strictly increasing affine reparametrization of the temperature channel (Corollary~\ref{cor:quantile_invariance}), so the routing partition is scale-free. The diagnostic entropy $\Psi$ is not affine-invariant, and its empirical direction with respect to correctness is opposite to that of $u$, with
$\mathrm{Pearson}(u,\mathrm{corr}) = +0.4763 \pm 0.0291,$ and
$\mathrm{Pearson}(\Psi,\mathrm{corr}) = -0.4301.$
The two quantities serve structurally different roles. $u$ is the operational quantity, and its routing partition does not change under affine reparameterization. $\Psi$ is a diagnostic, and its reading depends on the scale.

The variance of $\Psi$ contracts relative to that of $u$ over the pooled RoIs. The empirical statistics are listed in Table~\ref{tab:routing_stats}.
\begin{table}[t]
\centering
\caption{Empirical statistics of the routing scalar, the diagnostic entropy, and the acquisition product, pooled over $n=5.4\times10^{5}$ RoIs and ten seeds on FLIR ADAS.}
\label{tab:routing_stats}
\small
\begin{tabular}{lccc}
\toprule
 & mean & std & var \\
\midrule
$u$ & 0.3673 & 0.2175 & 0.047301 \\
$\Psi$ & 0.4982 & 0.0204 & 0.000416 \\
$\Delta$ & 0.1827 & 0.1086 & 0.011792 \\
\bottomrule
\end{tabular}
\end{table}
The variance ratio and the derived degeneracy parameter are
$\frac{\mathrm{Var}(\Psi)}{\mathrm{Var}(u)} = 8.8\times 10^{-3},$
$\frac{\mathrm{std}(\Psi)}{\mathrm{std}(u)} = 9.4\times 10^{-2}, $ and $\eta = 6.5\times 10^{-3}.$
The linear fit $\Delta_{\mathrm{acq}}=a u + b$ gives
$a = 0.4982,$ $ b = -0.0003, $
$R^{2} = 0.9958$, and $ \sigma_{\mathrm{res}} = 0.007054.$
The slope $a$ coincides with $\bar{\Psi}=0.4982$ to four decimal places, exactly as predicted by Proposition~\ref{prop:finite-S-degeneracy}. Corollary~\ref{cor:acq_degeneration} predicts $R^{2}\geq 1-\eta = 0.9936$; the measured $R^{2}=0.9958$ confirms the bound. The acquisition product $\Delta_{\mathrm{acq}}$ is therefore a rescaled copy of $u$ in this regime, and $\Psi$ contributes no independent ranking information.

A partial-correlation control addresses the possibility that the routing value of $u$ is a box-scale artifact. Because $u$ is negatively correlated with box area (Spearman $\rho(u,\mathrm{area})=-0.36$), box scale is a potential confound. Controlling for area leaves the $u$-correctness association essentially unchanged. The partial correlations are $\mathrm{Pearson}(u,\mathrm{corr}\mid\mathrm{area}) = +0.4677$ and $\mathrm{Pearson}(\Psi,\mathrm{corr}\mid\mathrm{area}) = -0.4284$, retaining $98\%$ and $99.6\%$ of the raw correlations respectively. The class confidence is weaker and also retains its sign, $\mathrm{Pearson}(\mathrm{conf},\mathrm{corr}) = -0.2144$ and $\mathrm{Pearson}(\mathrm{conf},\mathrm{corr}\mid\mathrm{area}) = -0.2048$. The routing value of $u$ and the opposite diagnostic direction of $\Psi$ are therefore structural properties of the evidence distribution, not box-scale artifacts.

Binning the pooled RoIs by $u$ reveals a three-state structure, not a monotone continuum (Table~\ref{tab:three_state}). The routing scalar therefore acts as a two-ended separator.

\begin{table}[t]
\centering
\caption{Three-state structure of the routing scalar $u$ on FLIR ADAS, pooled over ten seeds. The $u_{\mathrm{med}}$ values of the four regimes are $0.263$--$0.271$ (bins 0--6), $0.2718$ (bin 7), $0.5620$ (bin 8), and $0.9943$ (bin 9). Regime labels describe the $u$-stratified structure. Router branches follow Eq.~\eqref{eq:quantile_routing} with pooled ten-seed thresholds $Q_{0.3}=0.267$ and $Q_{0.9}=0.714$. Bins 0--6 are grouped as a single background regime. Their median $u$ spans $Q_{0.3}$, so some samples within this regime route to pseudo. The saturated regime is the only regime where mean IoU decreases relative to the preceding bin; this drop is what makes $u$ a two-ended separator and not a monotone ranking.}
\label{tab:three_state}
\small
\resizebox{\columnwidth}{!}{%
\begin{tabular}{ccccc}
\toprule
regime & $u$ range & $\mathrm{frac}(\mathrm{IoU}>0.5)$ & mean IoU & router branch \\
\midrule
background & bins 0--6 & $\approx 0$ & $\approx 0$ & discard \\
transition & bin 7 & $0.0948$ & $0.0861$ & pseudo \\
detection & bin 8 & $0.9274$ & $0.7652$ & pseudo \\
saturated & bin 9 & $0.3184$ & $0.2895$ & annotate \\
\bottomrule
\end{tabular}%
}
\end{table}

The quantile router of Eq.~\eqref{eq:quantile_routing} distinguishes the two ends of the distribution and does not need monotonicity in between. The discard branch at the bottom $30\%$ of $u$ has mean error $0.9941 \pm 0.0063$, and the annotate branch at the top $10\%$ has mean error $0.6773 \pm 0.0583$. We retain raw $u$ for routing to keep the router aligned with the theoretical treatment of Section~\ref{sec:routing}. The diagnostic entropy $\Psi$ remains outside the routing rule. Remark~\ref{rem:q1_validation} finds that $u$ and $H(\hat{p})$ are weakly negatively associated, consistent with the opposite signs of the $u$-correctness and $\Psi$-correctness correlations above. The detection regime (bin 8) has the highest mean IoU yet falls in the pseudo branch, while the annotate branch captures the saturated bin 9. The two-ended separation routes by distance from the bulk of the evidence distribution, consistent with Remark~\ref{rem:q1_validation}. A large $S$ need not indicate certainty, and the annotate branch is populated by ambiguous rather than clean detections. In summary, $u$ is a scale-invariant routing score, and its effectiveness rests on two-ended separation. $\Psi$ is a scale-dependent diagnostic, and its ranking contribution is absorbed by $u$ through the finite-evidence degeneracy of Proposition~\ref{prop:finite-S-degeneracy} and Corollary~\ref{cor:acq_degeneration}. The fixed-threshold router yields a lower per-image switch fraction than the quantile router during training (Appendix~\ref{app:switch_rate}), so affine invariance of the partition does not guarantee greater routing stability.

\subsection{Design Extensions: Replay and Distillation}
\label{sec:replay_distill}

We sketch replay and distillation extensions to illustrate how the same scalar \(u\) and \(|\Delta T|\) can be reused. Empirical evaluation is left to future work. Fig.~\ref{fig:lifecycle}(d) marks them as dashed design-extension components.

\paragraph{Gradient-conflict replay.} Let $g_i = \nabla_\theta \mathcal{L}(\theta; x_i)$ be the gradient of a stored sample and $g_{\text{cur}}$ the current batch gradient. The conflict score is
\begin{equation}
c_i = 1 - \cos(g_i, g_{\text{cur}}).
\label{eq:conflict}
\end{equation}
The temperature-weighted score is
\begin{equation}
c'_i = c_i (1 + \lambda |\Delta T_i|),
\label{eq:tempweight}
\end{equation}
and each stored prototype is a triple $(\mu_i, \sigma_i^2, \Delta T_i)$, so memory drops from $O(NHW)$ to $O(Nd)$ with $d \ll HW$. We fix $N = 500$ throughout.

\paragraph{Frequency-decoupled distillation.} Let $F_L$ and $F_H$ be low- and high-frequency components from a Discrete Cosine Transform (DCT) partition. The high-frequency weight is
\begin{equation}
w_{\text{high}}(x) = w_{\text{high}}^0 \bigl(1 + \eta \tanh(\Delta T_x / T_{\text{ref}})\bigr).
\label{eq:whigh}
\end{equation}
When the target is thermally distinct, high-frequency infrared edges are reliable and merit strict supervision. When it is not, high-frequency content is dominated by sensor noise. This contrasts with FreqKD~\cite{thaker2026freqkd}, which applies a globally fixed high-frequency weight. TEL makes the weight a function of the physical contrast $\Delta T$, so the supervision strength adapts per sample.

The uncertainty-weighted distillation loss is
\begin{equation}
\begin{aligned}
\lefteqn{\mathcal{L}_{\text{distill}} = \sum_i (1 - u_i)^\beta \Bigl[
 \lambda_L \|F_L^{\text{RGB}} - F_L^{\text{IR}}\|^2 }\\
& + \lambda_H w_{\text{high}}(x_i) \log\bigl(1 + \|F_H^{\text{RGB}} - F_H^{\text{IR}}\|^2\bigr) \Bigr].
\end{aligned}
\label{eq:distillloss}
\end{equation}
For samples drawn from prototype memory, the distillation weight is multiplied by $1 + \zeta c'_i$.

\begin{remark}[Monotone weighting in replay and distillation]
\label{rem:weighting_consistency}
(i) Consider the entropy-regularized replay selection problem
\begin{equation}
\min_{w \ge 0} \;\; \sum_i w_i \cdot \mathrm{drift}_i + \tau \sum_i w_i \log w_i \quad \text{s.t.} \quad \sum_i w_i = 1,
\label{eq:replayopt}
\end{equation}
where $\tau > 0$ and $\mathrm{drift}_i \propto 1/(1 + \lambda|\Delta T_i|)$. The optimal weights are
\begin{equation}
w_i^* = \frac{\exp(-\mathrm{drift}_i/\tau)}{\sum_j \exp(-\mathrm{drift}_j/\tau)},
\label{eq:optimalweight}
\end{equation}
and $w_i^*$ is strictly increasing in $|\Delta T_i|$.

(ii) Let $r(u)$ denote the teacher's reliability, assumed non-increasing in student uncertainty $u$. Let $w(u)$ be any non-increasing weighting with $\int w(u) p(u) du = 1$. Then
\begin{equation}
\int w(u)\, r(u)\, p(u) \, du \;\ge\; \int r(u)\, p(u) \, du.
\label{eq:consistency}
\end{equation}
In particular, $w(u) = (1-u)^\beta$ with $\beta > 0$ satisfies the condition.
\end{remark}

\begin{proof}[Proof]
See Appendix~\ref{app:proofs_replay_distill}.
\end{proof}

\subsection{Optimization Layer: Closed-Loop Convergence}
\label{sec:closedloop}
 Algorithm~\ref{alg:tel} summarizes the full procedure. Each step is the numerical evaluation of the components derived above. Step~S2 implements the dual-channel decoding of Fig.~\ref{fig:lifecycle}(a). The multi-scale contrast of Eq.~\eqref{eq:multiscale}, the modulation of Eq.~\eqref{eq:multiscale_mod}, the augmented and modulated evidence of Eqs.~\eqref{eq:temp_prior} and~\eqref{eq:evidence}, as well as the cognitive uncertainty $u$ are computed in Step~S3. Step~S4 applies the router of Eq.~\eqref{eq:routing}. The temperature-weighted score of Eq.~\eqref{eq:tempweight} together with the replay set is evaluated in Step~S6, and Step~S7 performs the gradient step of Eq.~\eqref{eq:unified}.

The convergence of this loop is governed by the routing switch rate. The routing switch rate of a trajectory is
\begin{equation}
\nu_T = \sum_{t=1}^{T-1} \mathbb{1}\bigl[\rho_{\theta^{(t)}} \neq \rho_{\theta^{(t+1)}}\bigr],
\label{eq:switch_rate}
\end{equation}
where $\rho_{\theta^{(t)}} \neq \rho_{\theta^{(t+1)}}$ means at least one sample changes its routing branch between consecutive iterations. The routing function is $u(x;\theta)=K/S(x;\theta)$.

\begin{lemma}[Lipschitz router: switch-rate bound]
\label{lem:switch_rate}
Assume: (i) $u(x;\theta)$ is $L_u$-Lipschitz in $\theta$; (ii) the parameter update satisfies $\|\theta_{t+1}-\theta_t\| \le \eta_t G$; (iii) in a neighborhood of each threshold $\tau \in \{\tau_{\text{high}}, \tau_{\text{low}}\}$, the density of $u$ is bounded by $p_\tau \le C$. Then
\begin{equation}
\mathbb{E}[\nu_T] \;\le\; 2n C L_u G \sum_{t=1}^{T-1}\eta_t.
\label{eq:switch_lipschitz}
\end{equation}
With $\eta_t = \eta_0/\sqrt{T}$ this gives $\mathbb{E}[\nu_T] = O(\sqrt{T})$.
\end{lemma}

\begin{proof}
See Appendix~\ref{app:proofs_convergence}.
\end{proof}

\begin{lemma}[Contraction router: stronger switch-rate bound]
\label{lem:contraction_rate}
If additionally the routing map is a contraction: there exists $\kappa \in (0,1)$ with $\|u(\cdot;\theta) - u(\cdot;\theta')\|_\infty \le \kappa\|\theta - \theta'\|$, and $\theta_t \to \theta^*$, then
\begin{equation}
\mathbb{E}[\nu_T] = O(\log T).
\label{eq:switch_contraction}
\end{equation}
\end{lemma}

\begin{proof}
See Appendix~\ref{app:proofs_convergence}.
\end{proof}

\begin{proposition}[Convergence with controlled routing switches]
\label{prop:convergence}
Fix the routing thresholds. Assume \(F\) is \(L\)-smooth within each
routing region, with bounded gradient variance \(\sigma^2\), bounded
gradients \(G\), and maximum loss discontinuity \(\Delta F\) at a
routing boundary. Let \(\nu_T\) be the iteration-level switch count
defined in Eq.~\eqref{eq:switch_rate}. With step size
\(\eta_t = \eta_0/\sqrt{T}\),
\begin{equation}
\begin{aligned}
\min_{t \le T} \mathbb{E}\bigl[\|\nabla F(\theta^{(t)})\|^2\bigr]
\;\leq\;&
\frac{F(\theta^{(0)}) - F^*}{\eta_0\sqrt{T}}
+ \frac{L\eta_0(\sigma^2 + G^2)}{2\sqrt{T}} \\
&+ \frac{\Delta F\,\mathbb{E}[\nu_T]}{\eta_0\sqrt{T}}.
\end{aligned}
\label{eq:switch_bound}
\end{equation}
Consequently:
\begin{itemize}[leftmargin=1.5em,itemsep=1pt]
\item If \(\mathbb{E}[\nu_T]=O(1)\), the switching term is
\(O(1/\sqrt{T})\).
\item If \(\mathbb{E}[\nu_T]=O(\log T)\) (the contraction case of
Lemma~\ref{lem:contraction_rate}), the switching term is
\(O(\log T/\sqrt{T})\).
\item If \(\mathbb{E}[\nu_T]=O(\sqrt{T})\) (the Lipschitz case of
Lemma~\ref{lem:switch_rate}), the switching term is \(O(1)\). 
\end{itemize}
\end{proposition}

\begin{proof}[Proof]
See Appendix~\ref{app:proofs_convergence}.
\end{proof}


\begin{algorithm}[t]
\caption{Temperature-Evidential Learning (TEL)}
\label{alg:tel}
\begin{algorithmic}[1]
\State \textbf{Input:} radiometric thermal streams, budget $\mathcal{C}$, thresholds $\tau_{\text{low}}, \tau_{\text{high}}$
\State \textbf{Round 0 (cold start):} Initialize from a COCO-pretrained YOLOv8-m teacher. Run $N_0$ warm-up iterations with $\gamma=0$, then anneal $\gamma$ linearly to its target over $N_1$ iterations.
\For{round $N = 1, 2, \dots$}
    \State \textbf{S1:} Acquire radiometric thermal images from the current domain.
    \State \textbf{S2:} Dual-channel decode: $G$ and $T$ (Fig.~\ref{fig:lifecycle}(a)).
    \State \textbf{S3:} Compute $\Delta T^{(s)}$ (Eq.~\eqref{eq:multiscale}), $m(\Delta T)$ (Eq.~\eqref{eq:multiscale_mod}), $e_{\text{aug},k}$ (Eq.~\eqref{eq:temp_prior}), $e_{\text{mod},k}$ (Eq.~\eqref{eq:evidence}), and $u$.
    \State \textbf{S4:} Route by Eq.~\eqref{eq:routing} or the quantile router of Eq.~\eqref{eq:quantile_routing}.
    \State \textbf{S5:} Annotators correct the annotation queue.
    \State \textbf{S6:} Compute $c'_i$ (Eq.~\eqref{eq:tempweight}); select replay set $\mathcal{R}^{(t)}$.
    \State \textbf{S7:} Incremental fine-tuning by Eq.~\eqref{eq:unified}.
    \State \textbf{S8:} Deploy.
    \State \textbf{S9:} If stopping criterion not met, return to S1.
\EndFor
\end{algorithmic}
\end{algorithm}
\section{Experiments}
\label{sec:experiments}

\subsection{Attribution Design}
\label{sec:protocol}

The second stage decomposes the paired thermal channel into its structural components through a continuous attribution chain. It proceeds in four steps. (i)~Is the temperature channel consumed by the detector? (ii)~Does the detection gain come from the physical channel? (iii)~Which component of the physical channel drives the calibration cost and which drives the detection gain? (iv)~Does the two-axis structure hold across datasets?

\paragraph{Datasets.} We evaluate on two radiometric benchmarks: FLIR ADAS and FLAME~3~\cite{flame3_paper}. FLIR ADAS provides a calibrated temperature matrix and supports the main attribution experiments. FLAME~3 serves as an auxiliary validation set for the redundancy check, the joint pairing-scale condition, the detection-axis decomposition, and the cross-dataset reading of the two axes. Table~\ref{tab:datasets} lists the two datasets and their roles. Dataset scale, acquisition details, and noise calibration are in Appendix~\ref{app:datasets}. The grayscale channel \(G\) is the camera's native 8-bit non-radiometric rendering, derived from its internal ISP. Although the visible and thermal sensors in FLIR ADAS are not frame-synchronized, this native grayscale channel is strictly pixel-aligned with the 14-bit calibrated TIFF data \(T\). We therefore adopt it as the grayscale channel, following the camera's actual tone-mapping and avoiding synthetic assumptions. We define a continuous physical drift coordinate \(s\) as the RGB mean intensity, which proxies ambient illumination, and stratify the FLIR ADAS images by quintiles of \(s\) to analyze \(P(y \mid s)\) and \(P(\Delta T \mid y, s)\) along this coordinate (Fig.~\ref{fig:deltaT_physics}(b)).

\begin{table}[t]
\centering
\caption{Datasets used in this work.}
\label{tab:datasets}
\small
\resizebox{\columnwidth}{!}{%
\begin{tabular}{lccc}
\toprule
Dataset & Modality & Calibrated $T$? & Role \\
\midrule
FLIR ADAS & RGB + 14-bit TIFF & \checkmark & Primary \\
FLAME~3 & Radiometric Long-Wave Infrared (LWIR) & \checkmark & Auxiliary \\
\bottomrule
\end{tabular}%
}
\end{table}

\paragraph{Two-stage oracle-free inference and training cache.}
All evaluations adopt a two-stage oracle-free protocol. Stage~1 uses a
fixed none seed-0 model to propose boxes; the same model generates the
training cache. Stage~2 constructs the $\Delta T$ and grayscale
pseudo-$\Delta T$ maps from Stage~1 boxes and performs a second forward
pass with modulation. Ground-truth boxes serve only for scoring and
are excluded from $\Delta T$ map construction.

\paragraph{Metrics.}
Detection metrics are mAP@0.5, Recall, and low-contrast recall (LCR). Uncertainty uses Expected Calibration Error (ECE)~\cite{guo2017calibration}. Theory verification additionally draws on the KL separability index $\Lambda(b)$ from Definition~\ref{def:kl_index}, the IPW risk $\hat{L}_{\mathrm{IPW}}$ from Eq.~\eqref{eq:ipw}, the routing switch statistics of Eq.~\eqref{eq:switch_rate} (Appendix~\ref{app:switch_rate}), and the discard error $\varepsilon_{\text{discard}}(\tau_{\text{low}})$ from Corollary~\ref{cor:discard_monotone}.

\paragraph{Definition of ECE and its scope.}
\label{sec:metrics_scope}
Because ECE is the primary arbiter of the calibration-cost claim, we state its definition explicitly. For a set of $N$ detections, we partition the per-RoI class-confidence scores into $M=15$ equally spaced bins $\{B_m\}_{m=1}^{M}$ and compute
\begin{equation}
\mathrm{ECE} = \sum_{m=1}^{M} \frac{|B_m|}{N} \left| \mathrm{acc}(B_m) - \mathrm{conf}(B_m) \right|,
\label{eq:ece_def}
\end{equation}
where $\mathrm{conf}(B_m)$ is the mean class confidence in bin $B_m$ and $\mathrm{acc}(B_m)$ is the empirical correctness rate in that bin, with correctness determined by an IoU threshold of $0.5$ against the ground-truth box. This is a classification-style ECE over per-RoI class confidence. It is the same definition for every attribution arm.

\paragraph{Implementation.} TEL is implemented in PyTorch and trained on 2$\times$RTX 3090 GPUs. The detector backbone is YOLOv8-m with an input resolution of $640\times640$. Full hyperparameters are in Appendix~\ref{app:datasets}. All main results are mean $\pm$ std over ten seeds on FLIR ADAS and eight seeds on FLAME~3, under the two-stage oracle-free protocol.

\paragraph{Cold start and initialization.}
TEL requires candidate boxes to compute $\Delta T$, while the resulting $u$ affects the next round's boxes. This dependency is resolved by the warm-up and $\gamma$-annealing schedule in Algorithm~\ref{alg:tel}, and exact step counts are given in Appendix~\ref{app:datasets}.

\paragraph{Temperature-aware augmentation.} Geometric augmentations (random crop, horizontal flip, multi-scale resize) are applied \emph{synchronously} to the grayscale image $G$ and the temperature matrix $T$, using the same affine transform. Photometric augmentations (color jitter, grayscale conversion) are applied only to $G$. The temperature matrix is never perturbed by photometric transforms. Mosaic augmentation is disabled: pasting non-adjacent crops creates artificial boundaries where the annular background $R_{\text{bg}}(b)$ no longer reflects the actual thermal context of the target. The dilation factor $r$ in Eq.~\eqref{eq:annulus} is defined relative to the box's current dimensions after augmentation, so the annulus scales with the target.

\subsection{Statistical Methodology and Confirmatory Family}
\label{sec:stat_method}

We evaluate the two-axis attribution under a three-tier family with Holm correction at the family-wise level of $0.05$. The Wilcoxon signed-rank test is the primary test; the paired $t$-test is also included in the tables.

\paragraph{Primary family ($m=2$).}
The primary family tests two headline calibration-axis claims under Holm correction at family-wise level $0.05$ with $m=2$, where the smallest threshold is $0.025$. Information-source advantage compares true-$\Delta T$ with scale-matched grayscale pseudo-$\Delta T$ on mAP@0.5 and is significant at Wilcoxon $p \approx 0.0098$. Joint-condition calibration cost compares true-$\Delta T$ with shuffled-$\Delta T$ on ECE and is significant at Wilcoxon $p=0.002$. Table~\ref{tab:paired} collects the paired contrasts.

\paragraph{Secondary family ($m=4$).}
The secondary family adds four additional calibration-axis probes under the same Holm correction, with family-wise level $0.05$ and $m=4$ giving a smallest threshold of $0.0125$. The true-vs-const-$\Delta T$ ECE contrast is significant at $p=0.002$ (Wilcoxon, $W=0$), which substantiates the joint-condition claim; breaking the absolute calibrated scale alone returns ECE to the broken-factor level. The remaining true-vs-pseudo contrasts on ECE, Recall, and LCR are not significant.

\paragraph{Detection-axis attribution.}
The var-phase vs.\ true-$\Delta T$ contrast on mAP is a pre-specified attribution of the detection axis. It is evaluated as a dedicated Wilcoxon signed-rank test, consistent with its role as an axis probe. The var-phase vs.\ random-$\Delta T$ contrast is given alongside as the isolation of the power-spectrum contribution. The FLAME~3 counterparts on both mAP and ECE are paired Wilcoxon signed-rank tests over the same eight paired seeds. At $n=8$ the smallest attainable two-sided $p$ is $2/2^{8}\approx 0.0078$. We report the attained $p$ without applying a family correction, because the FLAME~3 contrasts test the cross-dataset reading of the two axes rather than constitute independent confirmatory claims.

\paragraph{Exploratory probes.}
Random modulation, true-vs-const-$\Delta T$ on Recall, and FLAME~3 exploratory contrasts are not part of the confirmatory family.

\begin{table}[t]
\centering
\caption{Paired contrasts on FLIR ADAS over ten seeds under the two-stage protocol. All Cohen's $d$ values are paired-effect sizes, computed from the per-seed paired differences. The primary family is true-vs-pseudo on mAP and true-vs-shuffled on ECE ($m=2$, Holm threshold $0.025$); the secondary family is true-vs-pseudo ECE/Recall/LCR and true-vs-const-$\Delta T$ ECE ($m=4$, Holm thresholds $0.0125$, $0.0167$, $0.025$). The detection-axis attribution (var-phase-vs-true mAP and ECE) is pre-specified and reported as a dedicated contrast pair. Rows below the third rule are exploratory. Positive $\Delta$ means the first arm is larger.}
\label{tab:paired}
\small
\setlength{\tabcolsep}{3pt}
\resizebox{\columnwidth}{!}{%
\begin{tabular}{llccccc}
\toprule
Contrast & Metric & $\Delta$ & $t$-test $p$ & Wilcoxon $p$ & Cohen's $d$ & Reading \\
\midrule
true $-$ pseudo & mAP & $+0.0221$ & $0.0077$ & $\mathbf{0.0098}$ ($W=3$) & $+1.08$ & primary, Wilcoxon \\
true $-$ shuffled & ECE & $+0.0882$ & $0.039$ & $\mathbf{0.002}$ ($W=0$) & $+0.76$ & primary, true higher \\
\midrule
true $-$ pseudo & ECE & $-0.0097$ & $0.77$ & $0.064$ & $-0.095$ & secondary, n.s. \\
true $-$ pseudo & Recall & $+0.0008$ & $0.66$ & n.s. & $+0.14$ & secondary, n.s. \\
true $-$ pseudo & LCR & $+0.0044$ & $0.60$ & n.s. & $+0.17$ & secondary, n.s. \\
true $-$ const-$\Delta T$ & ECE & $+0.0916$ & $0.019$ & $\mathbf{0.002}$ ($W=0$) & $+0.90$ & secondary, Holm-pass \\
\midrule
var-phase $-$ true & mAP & $+0.0185$ & $0.014$ & $0.014$ ($W=4$) & $+0.96$ & detection-axis \\
var-phase $-$ random & mAP & $-0.0156$ & $0.0022$ & $0.004$ ($W=1$) & $-1.34$ & spectrum contribution \\
var-phase $-$ true & ECE & $-0.0271$ & $0.47$ & $0.38$ ($W=21$) & $-0.24$ & calibration-axis, n.s. \\
\midrule
true $-$ random & mAP & $-0.0341$ & $<0.001$ & $0.002$ ($W=0$) & $-1.83$ & exploratory \\
true $-$ random & LCR & $+0.0008$ & $0.86$ & $0.77$ & $+0.06$ & exploratory, n.s. \\
true $-$ shuffled & mAP & $+0.0045$ & $0.44$ & $0.16$ & $+0.25$ & exploratory, n.s. \\
\bottomrule
\end{tabular}%
}
\vspace{2pt}
\par\small \par\small The primary family and the true-vs-const-$\Delta T$ ECE contrast are significant under Holm correction at $m=2$ and $m=4$ respectively. For the detection-axis pair, the var-phase-vs-true ECE contrast is not significant and is the FLIR ADAS side of the calibration-axis cross-dataset asymmetry (Table~\ref{tab:flame3}). Other contrasts are secondary or exploratory.
\end{table}

\subsection{Step 1: Is the Temperature Channel Consumed?}
\label{sec:step_consumed}

Table~\ref{tab:attribution_consumption} lists the consumption probe. The true-$\Delta T$ arm increases the modulated evidence magnitude $e_{\text{mod}}$ by $+38.42\%$ relative to zero modulation ($p=4.2\times10^{-25}$), and increases the top detection score by $+2.48\%$ ($p=8.7\times10^{-43}$). The random-modulation arm increases $e_{\text{mod}}$ by only $+8.78\%$ and leaves the score essentially unchanged ($+0.05\%$, $p=9.5\times10^{-3}$). The channel is consumed, and this consumption is dominated by the physical contrast as opposed to generic modulation.

\begin{table}[t]
\centering
\caption{Attribution-consumption probe on FLIR ADAS ($200$ images). The modulated evidence magnitude $e_{\text{mod}}$ and the top detection score are compared across zero modulation, true calibrated $\Delta T$, and random modulation.}
\label{tab:attribution_consumption}
\small
\setlength{\tabcolsep}{3pt}
\resizebox{\columnwidth}{!}{%
\begin{tabular}{lccc}
\toprule
Probe & $e_{\text{mod}}$ mean & $\Delta e_{\text{mod}}$ & Top score \\
\midrule
zero modulation & $0.002453$ & reference & $0.8647$ \\
true $\Delta T$ & $0.003395$ & $+38.42\%$ ($p=4.2\times10^{-25}$) & $0.8862$ ($+2.48\%$, $p=8.7\times10^{-43}$) \\
random modulation & $0.002668$ & $+8.78\%$ ($p=2.97\times10^{-2}$) & $0.8651$ ($+0.05\%$, $p=9.5\times10^{-3}$) \\
\bottomrule
\end{tabular}%
}
\end{table}

\subsection{Step 2: Does the Gain Come from the Physical Channel?}
\label{sec:step_physical}

The second step of the attribution chain isolates the information source. We compare true $\Delta T$ against a scale-matched grayscale pseudo-$\Delta T$ control under parameter and FLOP matching. The contrast isolates the information source (temperature vs.\ grayscale) at matched scale. Table~\ref{tab:deltaTb} provides the seven-arm comparison. The paired contrasts on FLIR ADAS (Table~\ref{tab:paired}) reveal two effects. The true-vs-pseudo contrast is significant (paired Wilcoxon $p=0.0098$, $W=3$, $d=+1.08$), indicating that the physical channel provides information lost in the collapsed grayscale rendering. The true-vs-random contrast is also significant, but with the opposite sign: random modulation achieves a higher mAP than true $\Delta T$. The detection-axis factor is isolated by the phase-randomized arm and analyzed in Step~3.

On FLAME~3, the auxiliary validation appears in Table~\ref{tab:flame3}. True $\Delta T$ is statistically indistinguishable from pseudo-$\Delta T$ on mAP, consistent with the high $\Delta G$/$\Delta T$ redundancy (Pearson $r=0.76$, Spearman $\rho=0.85$); relative to random modulation, true $\Delta T$ is significantly worse on mAP and ECE. The physical-attribution claim therefore rests on the FLIR true-versus-pseudo contrast. The FLAME~3 random-superiority result and the detection-axis decomposition on FLAME~3 are analyzed in Step~4.

\begin{table*}[t]
\centering
\caption{Overall comparison on FLIR ADAS (10-seed prediction, all bins pooled). The primary acceptance test is TEL(true $\Delta T$) versus Grayscale pseudo-$\Delta T$. }
\label{tab:deltaTb}
\small
\resizebox{\textwidth}{!}{%
\begin{tabular}{lccccc}
\toprule
Variant & mAP@0.5 & Recall & LCR & ECE $\downarrow$ & Params/FLOPs matched? \\
\midrule
Baseline (no modulation) & $0.4664 \pm 0.0000$ & $0.3761 \pm 0.0000$ & $0.4279 \pm 0.0000$ & $0.4585 \pm 0.0000$ & -- \\
Random modulation & $0.5190 \pm 0.0096$ & $0.3767 \pm 0.0028$ & $0.4247 \pm 0.0117$ & $0.0373 \pm 0.0068$ & \checkmark \\
Grayscale pseudo-$\Delta T$ (matched) & $0.4628 \pm 0.0135$ & $0.3750 \pm 0.0040$ & $0.4211 \pm 0.0067$ & $0.1512 \pm 0.0298$ & \checkmark \\
Shuffled true $\Delta T$ & $0.4804 \pm 0.0105$ & $0.3744 \pm 0.0040$ & $0.4228 \pm 0.0148$ & $0.0534 \pm 0.0135$ & \checkmark \\
const-$\Delta T$  & $0.4833 \pm 0.0140$ & $0.3789 \pm 0.0040$ & $0.4240 \pm 0.0142$ & $0.0500 \pm 0.0210$ & \checkmark \\
TEL (true $\Delta T$) & $0.4849 \pm 0.0152$ & $0.3758 \pm 0.0024$ & $0.4255 \pm 0.0136$ & $0.1415 \pm 0.1069$ & \checkmark \\
var-phase  & $0.5034 \pm 0.0063$ & $0.3754 \pm 0.0040$ & $0.4227 \pm 0.0117$ & $0.1144 \pm 0.0054$ & \checkmark \\
\bottomrule
\end{tabular}%
}
\vspace{2pt}
\par\small \par\small Random removes pairing, scale, spectrum, and phase; pseudo removes the physical channel; shuffled removes pairing; const-$\Delta T$ removes absolute scale; var-phase removes spatial phase only.
\end{table*}

\begin{table*}[t]
\centering
\caption{Auxiliary validation on FLAME~3 (radiometric wildfire detection, $|\Delta T| \approx 30$--$60^\circ$C). Mean $\pm$ std over eight seeds. All $p$ values are paired Wilcoxon signed-rank tests over the eight paired seeds, where the smallest attainable two-sided $p$ is $0.0078$.}
\label{tab:flame3}
\small
\begin{tabular}{lcccc}
\toprule
Variant & mAP@0.5 & Recall & Low-contrast Recall & ECE $\downarrow$ \\
\midrule
Random modulation & $0.9238 \pm 0.0053$ & $0.9023 \pm 0.0018$ & $0.9337 \pm 0.0048$ & $0.0421 \pm 0.0056$ \\
Grayscale pseudo-$\Delta T$ & $0.9120 \pm 0.0060$ & $0.8993 \pm 0.0013$ & $0.9281 \pm 0.0037$ & $0.3066 \pm 0.0124$ \\
TEL (true $\Delta T$) & $0.9161 \pm 0.0038$ & $0.9005 \pm 0.0025$ & $0.9300 \pm 0.0064$ & $0.3095 \pm 0.0230$ \\
const-$\Delta T$ & $0.9153 \pm 0.0074$ & $0.8988 \pm 0.0025$ & $0.9281 \pm 0.0073$ & $0.2319 \pm 0.0662$ \\
var-phase & $0.9225 \pm 0.0056$ & $0.9019 \pm 0.0025$ & $0.9342 \pm 0.0064$ & $0.0639 \pm 0.0058$ \\
\midrule
\multicolumn{5}{l}{true $-$ pseudo $= +0.0041$ mAP (paired Wilcoxon $p=0.082$); true $-$ random $= -0.0077$ ($p=0.0074$);}\\
\multicolumn{5}{l}{var-phase $-$ true $= +0.0064$ mAP ($p=0.078$, Cohen's $d=+0.74$); var-phase $-$ random $= -0.0013$ ($p=0.83$);}\\
\multicolumn{5}{l}{var-phase $-$ true $= -0.2456$ ECE ($p=0.0078$, the eight-seed floor, Cohen's $d=-9.04$);}\\
\multicolumn{5}{l}{ECE true $-$ random $= +0.2674$ ($p<0.001$); ECE true $-$ const-$\Delta T$ $= +0.0775$ ($p=0.016$).}\\
\bottomrule
\end{tabular}
\end{table*}

\subsection{Step 3: Attribution of the Calibration and Detection Axes}
\label{sec:step_which}

The attribution chain separates the calibration-axis and detection-axis effects of each arm. On the calibration axis, const-$\Delta T$ preserves within-image relative temperature structure but removes the absolute calibrated scale, and shuffled-$\Delta T$ destroys paired correspondence while preserving amplitude. A $2\times 2$ interaction on ECE (Table~\ref{tab:interaction}) indicates that paired correspondence and absolute calibrated scale act jointly on ECE (interaction contrast $+0.0756$, Wilcoxon $p=0.0059$). The joint condition drives the ECE cost. Preserving both factors raises ECE by about $3\times$ over the broken-factor arms, and the paired contrasts establish that breaking either factor alone is sufficient to return ECE to the broken-factor level. Both the true-vs-const and true-vs-shuffled contrasts yield $p=0.002$ ($W=0$), with $d=+0.90$ and $d=+0.76$ respectively. The cost is confined to the calibration axis.

Post-hoc temperature scaling absorbs more of the scale cost than of the pairing cost. Each arm is fitted with a single scalar $T$ on the validation set by minimizing the negative log-likelihood (NLL). The scalar scales the logit of the predictive probability $p_k=\alpha_k/S$. The resulting ECE is computed with $M=15$, the same bin count as Table~\ref{tab:deltaTb}. After recalibration, the true-vs-shuffled gap survives ($d=+0.0665$, $p=0.002$), whereas the true-vs-const gap is no longer significant ($d=+0.0494$, $p=0.16$). Roughly half of the const gap and about a quarter of the shuffled gap are removed by monotone rescaling (Table~\ref{tab:posthoc}). The two factors therefore differ in the proportion of their ECE cost that recalibration can undo, not in the kind of distributional change they induce. At $M=15$, the TV distance from true is comparable for shuffled ($0.42 \pm 0.20$) and const ($0.49 \pm 0.16$; paired Wilcoxon $p=0.11$), and both are far below the distance to random ($0.72 \pm 0.06$). Distributional distance does not separate the two factors, but the share absorbed by recalibration does.

Seed s2 isolates a non-rescalable component. Its fitted temperature is $T=3.89$, far above the $0.72$--$0.78$ range of the other true seeds, yet its ECE barely moves under recalibration ($0.4574 \to 0.4433$). A temperature of $3.89$ compresses the logit by nearly a factor of four, and ECE does not respond. The residual mis-calibration is therefore not absorbed by a monotone rescaling.  
\begin{table}[t]
\centering
\caption{Post-hoc temperature scaling on FLIR ADAS (10 seeds). $T$ is fitted per arm on the validation set with NLL as the objective; $\mathrm{ECE}_{\text{TS}}$ uses $M=15$. $T$ is the median across seeds; the true arm's mean is affected by seed s2 ($T=3.89$; Section~\ref{sec:step_which}). Raw ECE is reproduced from Table~\ref{tab:deltaTb}; $\Delta$ECE is computed from per-seed means before rounding.}
\label{tab:posthoc}
\small
\resizebox{\columnwidth}{!}{%
\begin{tabular}{lcccc}
\toprule
Variant & ECE (raw) & $T$ (median) & ECE$_{\text{TS}}$ & $\Delta$ECE \\
\midrule
Baseline (no modulation) & $0.4585$ & $2.645$ & $0.4588$ & $+0.0003$ \\
Random modulation & $0.0373$ & $0.807$ & $0.0166$ & $-0.0206$ \\
Grayscale pseudo-$\Delta T$ & $0.1512$ & $0.661$ & $0.0860$ & $-0.0652$ \\
Shuffled true $\Delta T$ & $0.0534$ & $0.706$ & $0.0172$ & $-0.0361$ \\
const-$\Delta T$ & $0.0500$ & $0.872$ & $0.0342$ & $-0.0156$ \\
TEL (true $\Delta T$) & $0.1415$ & $0.763$ & $0.0837$ & $-0.0578$ \\
var-phase & $0.1144$ & $0.707$ & $0.0396$ & $-0.0747$ \\
\bottomrule
\end{tabular}%
}
\end{table}

The detection axis responds to a different factor. The phase-randomized arm preserves per-sample variance and the power-spectrum magnitude and removes only the spatial phase. It reaches mAP $0.5034 \pm 0.0063$ over true $\Delta T$ ($0.4849 \pm 0.0152$; paired Wilcoxon $p=0.014$, $W=4$, $d=+0.96$) with 52k predictions against true's 54k, at the same prediction count. Flattening the power spectrum adds a further $+0.0156$ ($p=0.004$, $d=-1.34$). Random's total advantage over true is $+0.0341$. The detection gain therefore follows structural de-structuring, not the spatial thermal structure.

\begin{figure}[t]
\centering
\includegraphics[width=\columnwidth]{"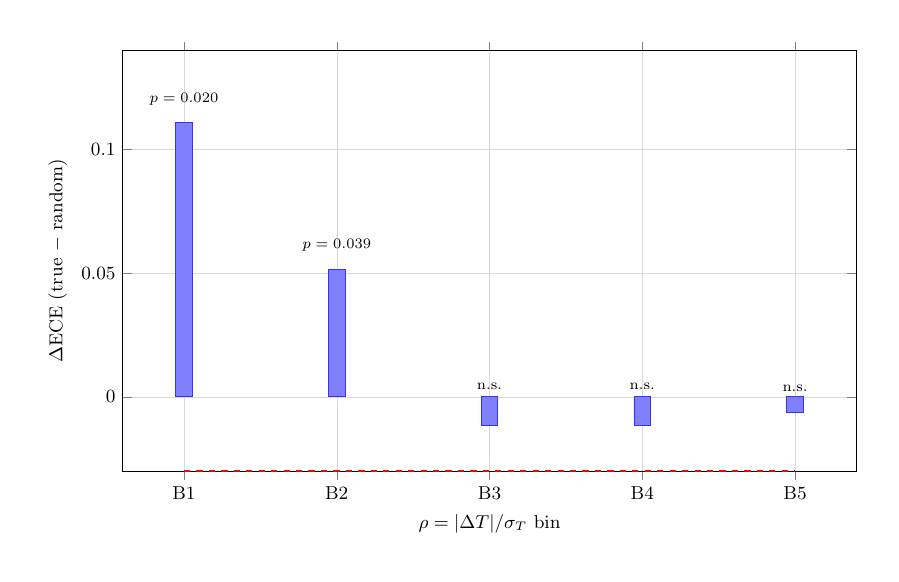"}
\caption{$\Delta$ECE (true $-$ random) versus per-box separability
ratio $\rho$ bins, with GT-derived quintile boundaries
$q_{20}/q_{40}/q_{60}/q_{80}=0.703/1.271/1.843/2.588$.}
\label{fig:rho_ece}
\end{figure}

The two axes are structurally orthogonal. On FLIR ADAS, the var-phase-vs-true contrast is significant on mAP ($p=0.014$) and insignificant on ECE ($p=0.47$, $W=21$, $d=-0.24$). The $\rho$-stratified analysis in Fig.~\ref{fig:rho_ece} localizes the calibration cost to the low-$\rho$ regime (B1 and B2), where it vanishes for $\rho > 1.27$ (B3--B5). The confidence-distribution diagnostic in Table~\ref{tab:confidence_dist} localizes random's low ECE to a global collapse of the confidence distribution (p90 $= 0.062$). The low ECE of random modulation originates from a global collapse of the confidence distribution, and does not arise from a sharper calibration of the joint condition. The calibration-axis attribution therefore rests on the joint-condition interaction, independent of any confidence-scale difference.

\begin{table}[t]
\centering
\caption{Confidence distribution over 10 seeds (paired protocol).}
\label{tab:confidence_dist}
\small
\resizebox{\columnwidth}{!}{%
\begin{tabular}{lccccc}
\toprule
Variant & paired & absolute & mean & p50 & p90 \\
\midrule
none & -- & -- & 0.493 & 0.500 & 0.500 \\
random & $\times$ & $\times$ & 0.083 & 0.032 & 0.062 \\
true & \checkmark & \checkmark & 0.222 & 0.110 & 0.637 \\
shuffled & $\times$ & \checkmark & 0.115 & 0.062 & 0.164 \\
const-$\Delta T$ & \checkmark & $\times$ & 0.158 & 0.059 & 0.652 \\
\bottomrule
\end{tabular}%
}
\vspace{2pt}
\par\small See Table~\ref{tab:interaction} for the 2$\times$2 interaction test.
\end{table}

\begin{table*}[t]
\centering
\caption{Joint-condition interaction test on ECE over ten paired seeds. The 2$\times$2 contrast $c = (\mathrm{ECE}_{1,1} - \mathrm{ECE}_{1,0}) - (\mathrm{ECE}_{0,1} - \mathrm{ECE}_{0,0})$ is positive and significant under the Wilcoxon signed-rank test. }
\label{tab:interaction}
\small
\resizebox{\textwidth}{!}{%
\begin{tabular}{lccc}
\toprule
Quantity & Value & 95\% CI & $p$ \\
\midrule
Interaction contrast $\bar{c}$ & $+0.0756$ & $[+0.011, +0.140]$ & $t$-test $0.0475$; Wilcoxon $0.0059$ \\
Cohen's $d$ & $+0.73$ & --- & --- \\
Leave-one-out $\bar{c}$ (sans s2) & $+0.0437$ & --- & $< 0.005$ \\
\midrule
Component $\mathrm{ECE}_{\text{true}} - \mathrm{ECE}_{\text{const}}$ & $+0.0916$ & --- & --- \\
Component $\mathrm{ECE}_{\text{shuffled}} - \mathrm{ECE}_{\text{random}}$ & $+0.0161$ & --- & --- \\
\bottomrule
\end{tabular}%
}
\vspace{2pt}
\par\small The interaction is carried by nine of ten seeds. The leave-one-out row gives the contrast without seed s2.
\end{table*}

\subsection{Step 4: Cross-Dataset Reading of the Two Axes}
\label{sec:step_second}
Table~\ref{tab:protocol} summarizes the discriminating power of each dataset. The FLAME~3 results are in Table~\ref{tab:flame3}.

The FLAME~3 detection-axis result carries the same sign and comparable magnitude but does not reach significance at eight seeds. The phase-randomized arm reaches mAP $0.9225$, above true $\Delta T$ ($0.9161$) and within noise of random ($0.9238$), while emitting 5178 predictions against true's 4641 and random's 7227. The paired contrast is $+0.0064$ ($p=0.078$, Cohen's $d=+0.74$), against $+0.0185$ on FLIR ADAS ($p=0.014$, $d=+0.96$). The detection-axis effect therefore has the same sign and comparable magnitude on both datasets, with the FLAME~3 estimate below the eight-seed significance floor.

The calibration axis is where the asymmetry emerges in the FLAME~3 results. var-phase reaches ECE $0.0639$ on FLAME~3, far below true $\Delta T$ ($0.3095$) and close to random ($0.0421$). The paired contrast is $-0.2456$ ($p=0.0078$, the smallest attainable at eight seeds, Cohen's $d=-9.04$). With respect to FLIR ADAS, var-phase and true $\Delta T$ share ECE of the same order ($0.1144$ vs.\ $0.1415$) with a paired contrast of $-0.0271$ that is not significant ($p=0.47$, Wilcoxon $p=0.38$, $d=-0.24$). The calibration-axis effect therefore registers on FLAME~3 but not on FLIR ADAS.

The same phase-randomized intervention produces a detection-axis effect of the same sign and comparable magnitude on both datasets, and a calibration-axis effect that manifests on one dataset and recedes on the other. The paired tests make the asymmetry explicit. The detection-axis effect is significant at ten FLIR seeds ($p=0.014$) and falls below the eight-seed floor on FLAME~3 ($p=0.078$). The same intervention is not significant as regards the calibration axis at ten FLIR seeds ($p=0.47$), yet it attains the eight-seed floor on FLAME~3 ($p=0.0078$). Each axis reaches significance on the dataset whose structure emphasizes it. The calibration axis responds to the dataset's structural contrast, but the detection axis does not.

\begin{figure}[htbp]
\centering
\includegraphics[width=\columnwidth]{"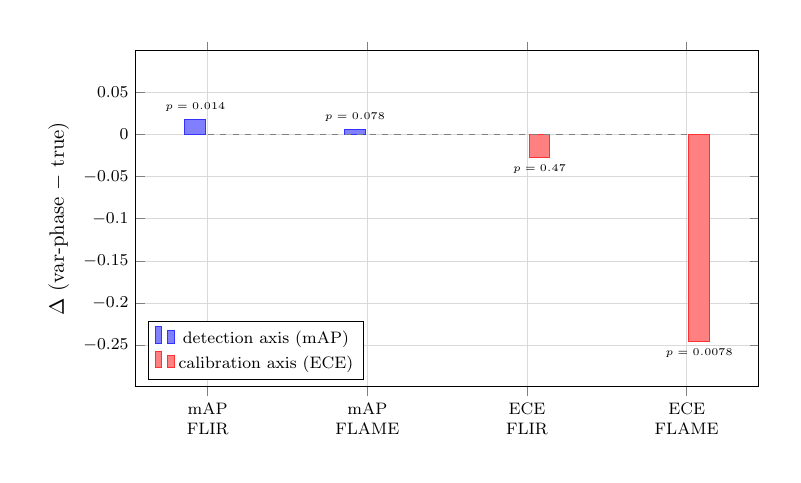"}
\caption{Cross-dataset reversal of the two axes. The same phase-randomized intervention (var-phase $-$ true) produces a detection-axis effect of the same sign and comparable magnitude on FLIR ADAS and FLAME~3, and a calibration-axis effect that is absent on FLIR ADAS ($p=0.47$) and significant on FLAME~3 ($p=0.0078$, the smallest attainable at eight seeds). Each axis is significant on the dataset where its structural counterpart is most pronounced, and insignificant on the other.}
\label{fig:two_axis_reversal}
\end{figure}

\begin{table*}[t]
\centering
\caption{Data conditions and their discriminating power. Each dataset
supplies a distinct validation axis.}
\label{tab:protocol}
\small
\resizebox{\textwidth}{!}{%
\begin{tabular}{lccc}
\toprule
Dataset & $\Delta T$ SNR & Pearson $r$ & Discriminates \\
\midrule
FLIR ADAS & channel SNR median $8.96$ (IQR $5.75$--$13.98$); per-box sep.\ SNR $0.69$ & low (collapse) & true vs.\ pseudo \\
FLAME~3 & $|\Delta T|$ p25/p50/p75 $=29.89/42.04/59.58^\circ$C & Pearson $r=0.76$ (redundant) & true vs.\ pseudo redundancy; random superiority check \\
\bottomrule
\end{tabular}%
}
\end{table*}

\begin{figure}[t]
    \centering
    \begin{subfigure}[b]{0.9\columnwidth}
        \centering
        \includegraphics[width=\textwidth]{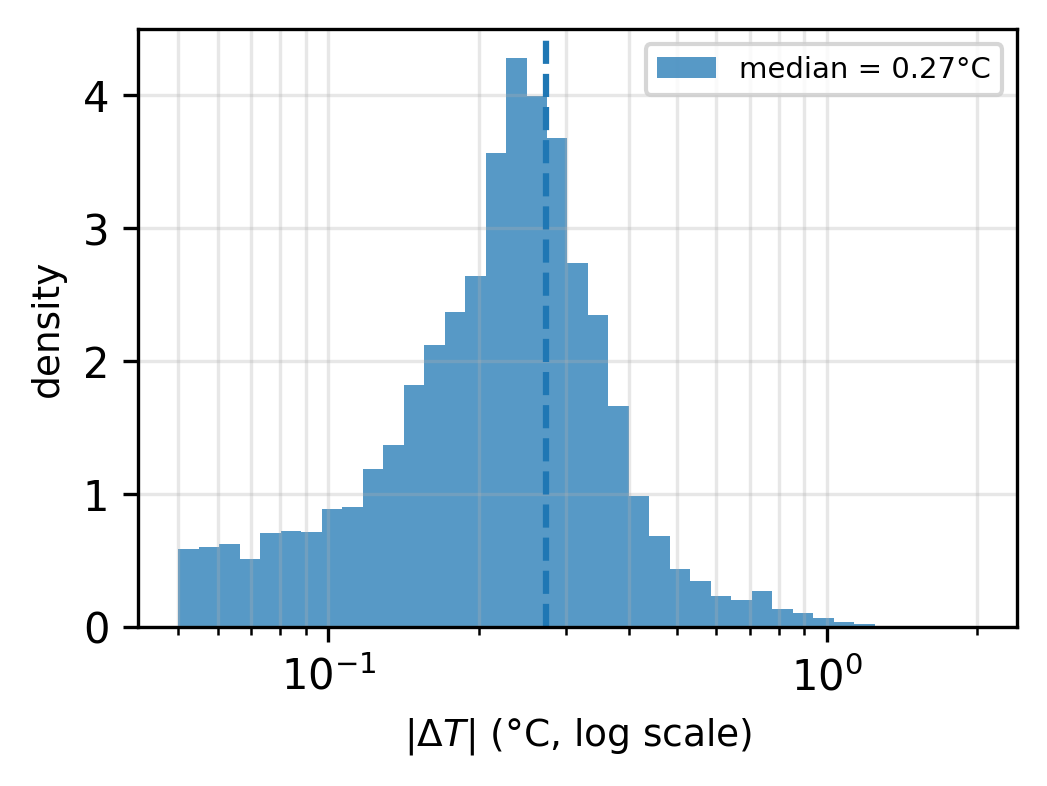}
        \caption{FLIR person $|\Delta T|$ ($n=12{,}267$)}
        \label{fig:cross_dataset_dt_a}
    \end{subfigure}
    \hfill
    \begin{subfigure}[b]{0.9\columnwidth}
        \centering
        \includegraphics[width=\textwidth]{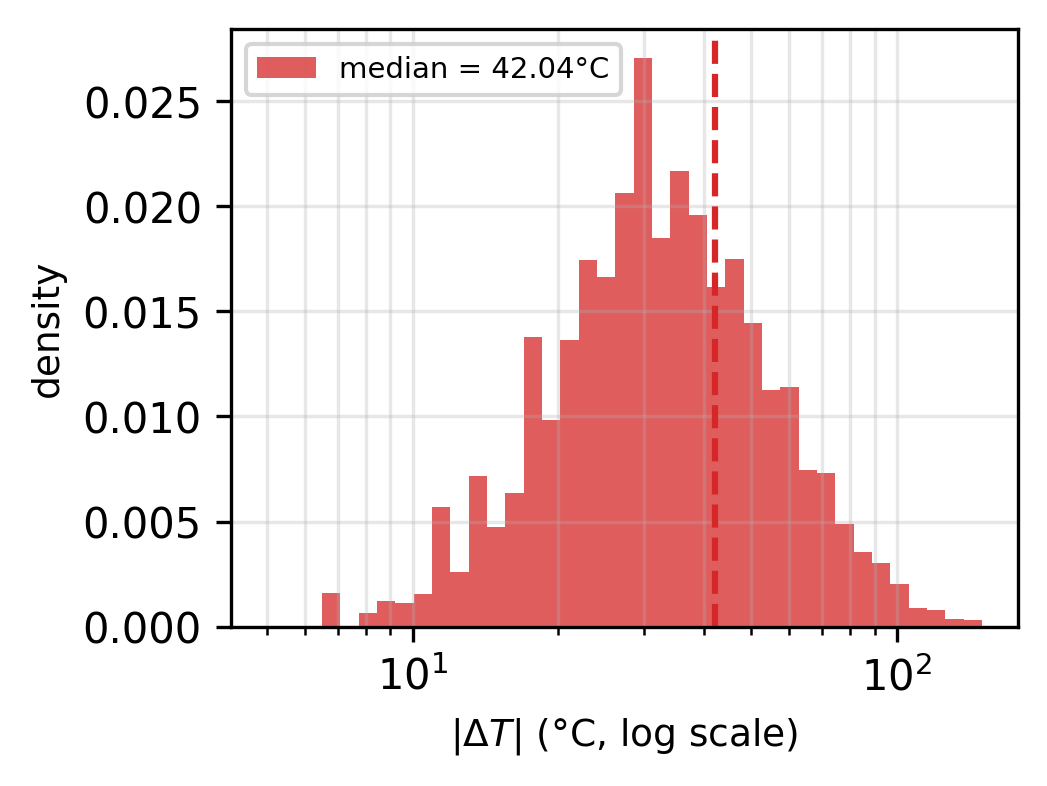}
        \caption{FLAME~3 fire $|\Delta T|$ ($n=2{,}126$)}
        \label{fig:cross_dataset_dt_b}
    \end{subfigure}
    \caption{Cross-dataset physical contrast.}
    \label{fig:cross_dataset_dt}
\end{figure}

\paragraph{Cross-dataset semantic consistency of physical thresholds.}
Step~4 has confirmed that the two-axis structure holds on a second dataset. A complementary question is what the retained absolute scale supplies at deployment time. Fig.~\ref{fig:cross_dataset_dt} indicates that the two datasets occupy non-overlapping physical regimes. FLIR person $|\Delta T|$ has median $0.27^\circ$C, while FLAME~3 fire $|\Delta T|$ has median $42.04^\circ$C. The separation is physical, not artefactual. On the target side, a clothed human is a passive source whose thermal contrast is attenuated by insulation. An open flame is an active source above $800$~K. The FLIR person-background contrast is compressed because both the insulated body surface and the near-ambient background sit close to ambient temperature (median $|\Delta T| = 0.27^\circ$C), but a FLAME~3 fire sits against a much colder background so the contrast is large (median $|\Delta T| = 42.04^\circ$C). Consequently the same numerical threshold $\tau$ in $^\circ$C selects physically different populations on the two datasets, and the difference is itself explained by the scene's thermal contrast rather than by a sensor artefact. The pass rates of four thresholds are detailed in Table~\ref{tab:cross_dataset_threshold}. In contrast, the gray threshold $\tau_G = 20$ on the native 8-bit rendering $G$ inverts across the two datasets ($73.6\%$ vs.\ $36.0\%$) and admits no physical interpretation.

\begin{table*}[!t]
\centering
\caption{Cross-dataset pass rates of physical and gray thresholds.}
\label{tab:cross_dataset_threshold}
\small
\resizebox{\textwidth}{!}{%
\begin{tabular}{lccc}
\toprule
Threshold $\tau$ & FLIR person pass & FLAME3 fire pass & Physical reading \\
\midrule
$0.5^\circ$C & $10.5\%$ & $99.95\%$ & FLIR: cold-background persons; FLAME~3: all fires \\
$5^\circ$C & $0\%$ & $99.9\%$ & FLIR: no person reaches $5^\circ$C; FLAME~3: most fires \\
$40^\circ$C & $0\%$ & $53.5\%$ & FLIR: no person reaches $40^\circ$C; FLAME~3: strong fires only \\
$\tau_G = 20$ (8-bit gray) & $73.6\%$ & $36.0\%$ & no physical reading \\
\bottomrule
\end{tabular}%
}
\par\small $\tau_G$ is on the native 8-bit rendering $G$ (0--255, dimensionless); it is not commensurable with the physical thresholds in $^\circ$C.
\end{table*}

\begin{remark}[Scale-invariant routing versus scale-semantic diagnosis]
\label{rem:scale_invariance_vs_semantics}
Corollary~\ref{cor:quantile_invariance} establishes that the routing partition is exactly invariant under any strictly increasing affine reparametrization $T' = aT + b$ with $a > 0$. The cross-dataset pass rates of Table~\ref{tab:cross_dataset_threshold} demonstrate that the physical reading of the same scale is not invariant and is in fact informative. The two statements are complementary. Routing is scale-invariant; the physical reading is scale-semantic.
\end{remark}

\subsection{Qualitative Analysis}
\label{sec:qualitative}

\begin{figure}[t]
\centering
\includegraphics[width=\columnwidth]{"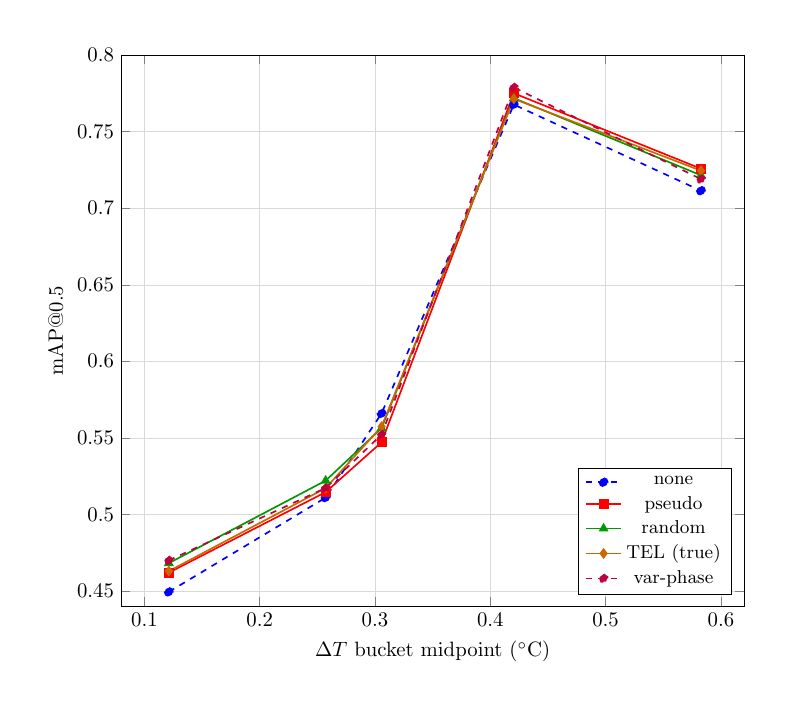"}
\caption{mAP@0.5 versus $\Delta T$ bucket midpoint on FLIR ADAS for five variants: none, pseudo, random, TEL (true), and var-phase. All five curves rise with $\Delta T$ up to B4, the high-transition bucket, and then decline at the redundant end (B5). The var-phase curve lies above true at B1, B2, and B4, below it at B3 and B5, with the largest positive gap at B4, the bucket where the injected-variance and true-variance scales are best matched. This is the qualitative counterpart of the paired $+0.0185$ mAP effect reported in Table~\ref{tab:paired}. Buckets are test-set GT $\Delta T$ median quintiles with boundaries $q_{20}/q_{40}/q_{60}/q_{80}=0.2424/0.2721/0.3392/0.5015^\circ$C.}
\label{fig:qualitative}
\end{figure}

Fig.~\ref{fig:qualitative} plots mAP@0.5 versus the $\Delta T$ bucket midpoint on FLIR ADAS for five variants: none, pseudo, random, TEL (true), and var-phase. All five curves rise with $\Delta T$ up to B4, the high-transition bucket, and then decline at the redundant end (B5). The var-phase gap over true is positive at B1, B2, and B4, negative at B3 and B5, and largest at B4. This is the qualitative counterpart of the paired $+0.0185$ mAP effect in Table~\ref{tab:paired}, and it shows that the detection-axis gain is not concentrated in a single $\Delta T$ regime.

\subsection{Discussion}
\label{sec:discussion}

\paragraph{Reading of the five-arm attribution.}
The five arms are controlled conditions. Const, shuffled, and var-phase each remove one factor, the random arm removes four with injected variance held matched, and the true arm retains all five. The five-arm table exposes two orthogonal axes with no shared driver. One responds to the joint pairing-scale condition, and the other is controlled by structural de-structuring. Neither axis's attribution depends on the confidence scale of the other. The paired tests on FLIR ADAS and FLAME~3 demonstrate the separation directly. Breaking either factor alone is sufficient to return ECE to the broken-factor level ($p=0.002$ for both true-vs-const and true-vs-shuffled), and the var-phase-vs-true ECE contrast is insignificant on the dataset where the pairing-scale structure dominates the calibration axis. The phase-randomized arm exceeds true $\Delta T$ while preserving per-sample variance and removing only spatial phase. This reverses the intuition that the calibrated channel should dominate any perturbation, and the five-arm design identifies why: the gain tracks  structural de-structuring, not the spatial thermal pattern. The same intervention is significant on FLIR ADAS and falls below the eight-seed floor on FLAME~3. TEL retains interpretability and attributable calibration. Random modulation achieves higher mAP but collapses the confidence distribution (Table~\ref{tab:confidence_dist}), so its ECE is not attributable to a physical condition.

\paragraph{Deployment value of the retained absolute scale.}
Physical thresholds in $^\circ$C exhibit cross-dataset semantic consistency, whereas gray thresholds invert without explanation. The absolute scale supplies physical traceability, cross-dataset semantic consistency, and deployment-time diagnosability. By Remark~\ref{rem:scale_invariance_vs_semantics}, this diagnostic value is complementary to the scale-invariant routing of Corollary~\ref{cor:quantile_invariance}. The spatial thermal structure carries no detection-gain contribution (main effect $-0.0185$). The absolute calibrated scale instead provides diagnostic value: the same threshold in $^\circ$C reads consistently across datasets.

\paragraph{Calibration and noise model.}
TEL assumes radiometric calibration. FLIR ADAS provides it, so $\Delta T$ is a physical quantity. If the camera outputs only relative counts, $\Delta T$ is defined up to an affine transform. By Corollary~\ref{cor:affine_invariance}, any strictly increasing affine reparametrization of $T$ preserves the direction of the uncertainty reduction and the ordering of $u$. By Corollary~\ref{cor:quantile_invariance}, the quantile router is exactly invariant, as verified empirically in Remark~\ref{rem:q3_validation}. The interpretation of $u$ as a physical quantity is then lost, and $T_{\text{ref}}$ in Eq.~\eqref{eq:whigh} must be set relative to the sensor's dynamic range. The empirical validation of the KL sandwich relies on the thermal camera's native 8-bit grayscale rendering to estimate the local grayscale ratio $x$. This rendering is strictly pixel-aligned with the 14-bit TIFF, but it introduces the camera's internal ISP non-linearities (e.g., tone-mapping). Nighttime infrared person detection also faces small target size, occlusion, low resolution, sensor non-uniformity, emissivity variation, registration error, clutter, and annotation noise.

\paragraph{Dataset scope and cross-dataset direction.}
FLIR ADAS and FLAME~3 differ by two orders of magnitude in $|\Delta T|$
and by more than a factor of three in the $\Delta G$--$\Delta T$ Pearson correlation,
which makes their attribution contrasts complementary
(Section~\ref{sec:step_second}, Tables~\ref{tab:protocol}
and~\ref{tab:flame3}). The dataset contrast modulates the calibration axis but leaves the detection axis invariant, which is the cross-dataset reading of the two-axis attribution. Beyond FLIR ADAS and FLAME~3, no public benchmark currently provides a calibrated temperature channel, and an annotation-budget sweep is left for future work.

\paragraph{Statistical stability.}
The joint condition drives a calibration cost that is confined to the calibration axis: preserving both factors raises ECE by about $3\times$ over the broken-factor arms, while detection accuracy is unaffected. De-structuring lowers ECE without being required for the mAP gain. The TEL ECE variance on FLIR ADAS is localized to the true-$\Delta T$ arm, where seed s2 exhibits a non-rescalable component (Section~\ref{sec:step_which}, Appendix~\ref{app:posthoc_details}). Shuffled and constant $\Delta T$ are stable. All main-text attribution results are reported at the nominal training scale \(\alpha=1.0\). A sweep over the modulation amplitude \(\alpha\) is provided in Appendix~\ref{app:alpha_sensitivity}.

Routing scalar and empirical probes. The routing layer is characterized by Proposition~\ref{prop:finite-S-degeneracy}, Corollary~\ref{cor:acq_degeneration}, and Section~\ref{sec:routing-scalar-structure}. The scalar $u$ is a two-ended separator rather than a monotone uncertainty ranking: its partial correlation with correctness controlling for box area is $+0.4677$. The diagnostic entropy $\Psi$ moves in the opposite direction ($\mathrm{Pearson}(\Psi,\mathrm{corr})=-0.4301$) and is equally robust to the area control ($-0.4284$). The degeneracy parameter $\eta=6.5\times 10^{-3}$ and the linear fit $\Delta_{\mathrm{acq}}=0.4982\,u-0.0003$ with $R^{2}=0.9958$ confirm the finite-evidence degeneracy predicted by Corollary~\ref{cor:acq_degeneration}: the acquisition product is a rescaled copy of $u$, and $\Psi$ contributes no independent ranking information in this regime. A pool-based top-$10\%$ selection probe on FLIR ADAS (5 seeds) does not improve over warmup ($0.5194\pm 0.0058$ vs.\ $u_{\mathrm{top10}}$ $0.4409\pm 0.0434$ and $H_{\mathrm{top10}}$ $0.4586\pm 0.0244$). The pool-based probe measures end-to-end annotation efficiency, whereas the routing-consistency results in Remarks~\ref{rem:q2b_validation}, \ref{rem:q2c_validation}, \ref{rem:q3_validation}, and Section~\ref{sec:routing-scalar-structure} characterize the local ordering behavior of $u$. For the paired contrasts in Tables~\ref{tab:paired} and~\ref{tab:deltaTb}, all attribution arms share the same evidential objective and modulation architecture, so any KL-regularization effect is common and cancels.
\section{Conclusion}
\label{sec:conclusion}

The calibrated thermal channel carries two structurally orthogonal axes. On the calibration axis, ECE tracks the joint pairing-scale condition, where paired correspondence and absolute calibrated scale interact, and breaking either factor alone returns ECE to the broken-factor level in raw ECE (Table~\ref{tab:paired}). Post-hoc recalibration absorbs more of the scale cost than of the pairing cost, so the two factors differ in proportion even though their distributional distances from the true arm are comparable. The detection axis follows structural de-structuring, not the spatial thermal pattern that calibration preserves. The phase-randomized arm raises mAP over true $\Delta T$ (Tables~\ref{tab:paired} and~\ref{tab:flame3}). Across the two datasets, the detection-axis effect holds in sign and comparable magnitude, while the calibration-axis effect registers on one and recedes on the other.

The same structural attribution may apply to any sensor that provides a calibrated physical channel paired with a visual one, such as LiDAR reflectance, radar cross-section, or multispectral bands. The joint condition is the causal driver behind one of the two structural components for the paired channel, and it is complemented by a structural-de-structuring component that governs detection. TEL implements the attribution testbed. It produces a single evidence-derived routing scalar $u$ and a complementary diagnostic entropy $\Psi$. It also proves a finite-evidence degeneracy theorem, which states that the asymptotic acquisition product degenerates to $u$ with the closed-form bound $R^{2}\geq 1-\eta$. Absolute scale provides deployment-time value through cross-dataset semantic consistency.

\balance

\appendix

\section{Proofs for Lemma~\ref{lem:sigmoid}, Proposition~\ref{prop:evidence_modulation}, and Corollary~\ref{cor:affine_invariance}}
\label{app:proofs_lemma}

\paragraph{Proof of Lemma~\ref{lem:sigmoid}.}
Let $m \in \mathcal{M}$ and define the log-odds transform $u(t) = \log(m(t)/(1-m(t)))$. Because $m$ is strictly monotone with $m(-\infty) = 0$ and $m(+\infty) = 1$, the map $u$ is a strictly increasing bijection, and $u'(t) = m'(t) / [m(t)(1-m(t))]$. The growth constraint becomes $u'(t) \leq 1$ for all $t$. If $m$ attains the constraint with equality everywhere, then $u'(t) = 1$, which integrates to $u(t) = t + c$ for some $c \in \mathbb{R}$. Inverting gives $m(t) = \sigma(t + c)$, which is Eq.~\eqref{eq:optmod} with $\kappa = 1$ and $\beta = c$. Rescaling the input argument by $\kappa \in (0,1]$ yields the family $\sigma(\kappa t + \beta)$, which satisfies the constraint strictly whenever $\kappa < 1$. Uniqueness follows from the Picard--Lindel\"of theorem applied to the logistic ODE $m' = m(1-m)$ with initial condition $m(0) = \sigma(\beta) \in (0,1)$. The deficit integrand is non-negative by the growth constraint; it is strictly positive on a set of positive measure unless the constraint holds with equality almost everywhere, which by continuity extends to everywhere.

\paragraph{Proof of Proposition~\ref{prop:evidence_modulation}.}
From the modulated evidence, $S_{\text{mod}} = S_{\text{aug}}[1 + \gamma m(\Delta T)a]$. Since $u_{\text{aug}} = K/S_{\text{aug}}$ and $u_{\text{mod}} = K/S_{\text{mod}}$, the uncertainty reduction follows as
$
u_{\text{aug}} - u_{\text{mod}}
= \frac{K}{S_{\text{aug}}} \cdot \frac{\gamma m a}{1 + \gamma m a},
$
which is Eq.~\eqref{eq:exact}. The two-sided bound follows from the monotonicity of $x \mapsto \gamma m x/(1+\gamma m x)$ on $[0,1]$ and the endpoint values $f(0) = 0$, $f(1) = \gamma m/(1+\gamma m)$. For the error-propagation bound, the same map has derivative $f'(x) = \gamma m/(1+\gamma m x)^2$, maximized at $x = 0$ with value $\gamma m$; the mean value theorem then gives $|f(\hat{a}) - f(a)| \le \gamma m \cdot \epsilon_a$.

\paragraph{Proof of Corollary~\ref{cor:affine_invariance}.}
By linearity of the annular mean, $\Delta T' = a \Delta T$. Since $a > 0$, $m'(\Delta T') = \sigma(\kappa a \Delta T + \beta)$ is a strictly increasing $C^1$ map with the same endpoints as $m$, hence belongs to $\mathcal{M}$. The map $m \mapsto \frac{\gamma m a}{1 + \gamma m a}$ in Eq.~\eqref{eq:exact} is strictly increasing, and $m'$ is comonotone with $m$ as a function of $\Delta T$; hence the direction of the uncertainty reduction is preserved. Finally, $u_{\text{mod}}$ is a decreasing function of $S_{\text{mod}}$, which is itself increasing in $m$; comonotonicity of $m, m'$ preserves the ordering of $u$ across samples.

\section{Proof of Proposition~\ref{prop:separability_sandwich}}
\label{app:proofs_separability}

\paragraph{Closed form and bounds.}
For two univariate Gaussians with equal variance \(\sigma^2\) and means \(0\) and \(\mu\), the KL divergence is
$
\KL(\mathcal{N}(\mu,\sigma^2)\|\mathcal{N}(0,\sigma^2)) = \frac{\mu^2}{2\sigma^2}.
$
Substituting \(\sigma^2 = \sigma_T^2 + s^{-2}\sigma_G^2\) gives Eq.~\eqref{eq:kl_closed}. Writing
$
x = \frac{s^{-2}\sigma_G^2}{\sigma_T^2} \ge 0,
$
we have
$
\Lambda(b)
= \frac12 \SNR_b^{\Delta T}\frac{1}{1+x}
= \frac12 \SNR_b^G.
$
The lower bound in Eq.~\eqref{eq:kl_sandwich} is the exact value with \(1/(1+x)\), and the upper bound \(\Lambda(b) \le \tfrac12 \SNR_b^{\Delta T}\) follows from \(1/(1+x)\le 1\). The lower bound is non-negative for all \(x\ge 0\). As \(x\to 0\), the two bounds coincide. As \(x\to \infty\), the upper bound stays fixed while the lower bound decays as \(1/(1+x)\), so the sandwich becomes loose.

\paragraph{Coupling to $u$.}
From Eq.~\eqref{eq:exact}, the reduction is
$
u_0 - u_{\text{mod}} = \frac{K}{S_0}\cdot\frac{\gamma m a}{1 + \gamma m a}.
$
In the small-modulation regime $\gamma m a \ll 1$, expand:
$
\frac{\gamma m a}{1+\gamma m a} = \gamma m a + O((\gamma m a)^2).
$
By Lemma~\ref{lem:sigmoid}, $m$ is a strictly increasing function of $\Delta T$; in the local Gaussian shift model, $1 - e^{-\Lambda(b)}$ is also a strictly increasing function of $\Delta T$, and both are $O(\Delta T^2)$ for small $\Delta T$. Hence to leading order $\gamma m a = 1 - e^{-\Lambda(b)} + O(\Lambda(b)^2)$ in the weak regime and $\to 1$ as $\Lambda(b) \to \infty$ in the strong regime. Combining the two asymptotics gives Eq.~\eqref{eq:u_kl_coupling}.

\section{Proof of Proposition~\ref{prop:routing_bound} and Corollary~\ref{cor:discard_monotone}}
\label{app:proofs_routing}

\paragraph{(i) Information-gain proxy and its large-evidence limit.}
Let \(p \sim \mathrm{Dir}(\alpha)\) with \(\alpha_k = e_k + 1\) and \(S = \sum_k \alpha_k\). The expected differential entropy of a Dirichlet-distributed categorical is
\[
\mathbb{E}_{p \sim \mathrm{Dir}(\alpha)}[H(p)]
= \psi(S + 1) - \frac{1}{S}\sum_{k=1}^{K} \alpha_k \psi(\alpha_k + 1),
\]
where \(\psi\) is the digamma function. Observing the true label \(y\) updates the posterior to \(\mathrm{Dir}(\alpha + e_y)\) with total strength \(S + 1\). The exact information gain is
\[
\mathrm{IG}(x)
= \mathbb{E}_{p \sim \mathrm{Dir}(\alpha)}[H(p)]
- \mathbb{E}_{y}\mathbb{E}_{p \sim \mathrm{Dir}(\alpha+e_y)}[H(p\mid y)].
\]
The exact information gain differs from this proxy. Expanding the digamma terms at large \(S\) gives
\[
\mathrm{IG}(x)
= \frac{K-1}{2S(x)} + O(S(x)^{-2})
= \frac{K-1}{2K}u(x) + O(S(x)^{-2}).
\]
Thus \(u(x)\) is strictly monotone in the leading term of the exact information gain, and \(\Delta_{\mathrm{acq}}(x)=u(x)\Psi(x)\) is used as a monotone acquisition proxy.

\paragraph{(ii) Pseudo-label error under correlation.}
Let $a = \Pr(f^{(t)}\text{ errs} \wedge f^{(t-1)}\text{ errs})$. The marginal error constraint gives $a + b = a + c = \varepsilon$, so $b = c$. The joint probability constraint gives $d = 1 - 2\varepsilon + a$, and probability non-negativity gives $a \in [\max(0, 2\varepsilon - 1), \varepsilon]$. Since $\rho = (a - \varepsilon^{2})/(\varepsilon(1 - \varepsilon))$, the range of $\rho$ follows. Under the acceptance rule, agreement requires both models to err or both to be correct; conditional on acceptance, the error probability is $a/(a+d)$. Substituting $a = \varepsilon^{2} + \rho\varepsilon(1-\varepsilon)$ and $d = 1 - 2\varepsilon + a$ gives the ratio in Eq.~\eqref{eq:quadratic}.

\paragraph{(iii) IPW generalization.}
Unbiasedness:
\[
\begin{aligned}
\mathbb{E}[\hat{L}_{\mathrm{IPW}}]
&= \frac{1}{n}\sum_{i=1}^{n}\mathbb{E}\!\left[
\frac{\mathbb{1}[\rho(x_i) = \text{annotate}]}{e(x_i)}\,\ell(f(x_i),y_i)
\right] \\
&= \frac{1}{n}\sum_{i=1}^{n}\mathbb{E}[\ell(f(x_i),y_i)]
= L_{\mathrm{test}}^{\mathcal A}.
\end{aligned}
\]
Boundedness of the summands by $1/e_{\min}$ gives, via Hoeffding,
\[
\Pr(|\hat{L}_{\mathrm{IPW}} - L_{\mathrm{test}}^{\mathcal A}| \ge t)
\le 2\exp(-2n e_{\min}^2 t^2).
\]
Setting $t = \sqrt{\log(2/\delta)/(2n e_{\min}^2)}$ yields the bound. Decomposing $L_{\text{test}} = \frac{n_A}{n} L_{\mathrm{test}}^{\mathcal A} + \frac{n_P}{n} L_{\mathrm{test}}^{\mathcal P} + \frac{n_D}{n} L_{\mathrm{test}}^{\mathcal D}$ and applying $L_{\mathrm{test}}^{\mathcal P} \le \varepsilon_{\text{pseudo}}$, $L_{\mathrm{test}}^{\mathcal D} \le \varepsilon_{\text{discard}}(\tau_{\text{low}})$ gives Eq.~\eqref{eq:router_gen}. The $\phi$-mixing extension replaces Hoeffding with the covariance-corrected inequality of the standard mixing concentration bound.

\paragraph{(iv) Online routing regret.}
Standard online gradient descent regret for convex $G$-Lipschitz losses gives Eq.~\eqref{eq:regret}.

\paragraph{Proof of Corollary~\ref{cor:discard_monotone}.}
Within the Dirichlet evidence model, the Bayes error of a $\mathrm{Dir}(\alpha)$-distributed categorical posterior with total strength $S$ satisfies
$
R^* \le \frac{K-1}{S+1}.
$
Substituting $S = K/u$ gives
$
R^* \le \frac{K-1}{K/u + 1}.
$
The right-hand side is strictly decreasing in $u$. Conditioning on $u < \tau_{\text{low}}$ and applying monotonicity gives
$
\varepsilon_{\text{discard}}(\tau_{\text{low}})
\le \frac{K-1}{K/\tau_{\text{low}} + 1}.
$
The right-hand side is strictly increasing in $\tau_{\text{low}}$.

\section{Proofs of Proposition~\ref{prop:finite-S-degeneracy} and Corollary~\ref{cor:acq_degeneration}}
\label{app:finite-S-degeneracy}

\paragraph{Proof of Proposition~\ref{prop:finite-S-degeneracy}.}
Write $\Psi=\bar{\Psi}+\delta$ with $\mathbb{E}[\delta]=0$. Then
$
\Delta_{\mathrm{acq}} = u\cdot\bar{\Psi} + u\cdot\delta.
$
The variance of the residual is bounded by
\[
\mathrm{Var}(u\cdot\delta)
\leq \mathbb{E}[u^{2}]\,\mathrm{Var}(\delta)
= \mathbb{E}[u^{2}]\,\mathrm{Var}(\Psi),
\]
using $\mathbb{E}[\delta]=0$ and the Cauchy--Schwarz inequality. Dividing by $\bar{\Psi}^{2}\,\mathrm{Var}(u)$ gives the definition of $\eta_S$. The pointwise bound follows from $|\Delta_{\mathrm{acq}}-\bar{\Psi}u| = |u\,\delta|$. The rank perturbation is of order $\eta_S^{1/2}$ by the variance bound. The $R^{2}$ bound follows from Corollary~\ref{cor:acq_degeneration}(iii).

\paragraph{Proof of Corollary~\ref{cor:acq_degeneration}.}
Part (i) is immediate from the decomposition. For part (ii), write $\Delta_{\mathrm{acq}}(x)=\bar{\Psi}u(x)(1+\varepsilon(x))$ with $|\varepsilon(x)|=|\delta(x)|/|\bar{\Psi}|\leq\varepsilon$. If $u(x)/u(y)>(1+\varepsilon)/(1-\varepsilon)$, then
$
\Delta_{\mathrm{acq}}(x)
= \bar{\Psi}u(x)(1+\varepsilon(x))
\geq \bar{\Psi}u(x)(1-\varepsilon)
> \bar{\Psi}u(y)(1+\varepsilon)
\geq \bar{\Psi}u(y)(1+\varepsilon(y))
= \Delta_{\mathrm{acq}}(y),
$
so the ordering is preserved. A flip requires the reverse inequality, hence $u(x)/u(y)\leq(1+\varepsilon)/(1-\varepsilon)$, i.e., relative $u$-gap at most $2\varepsilon/(1-\varepsilon)$. For part (iii), under independence of $\delta$ and $u$,
$
\mathrm{Cov}(\Delta_{\mathrm{acq}},u) = \bar{\Psi}\,\mathrm{Var}(u),
$
$
\mathrm{Var}(\Delta_{\mathrm{acq}})
= \bar{\Psi}^{2}\,\mathrm{Var}(u)
+ \mathbb{E}[u^{2}]\,\mathrm{Var}(\Psi),
$
so
\[
R^{2}
= \frac{\mathrm{Cov}(\Delta_{\mathrm{acq}},u)^{2}}
        {\mathrm{Var}(\Delta_{\mathrm{acq}})\,\mathrm{Var}(u)}
= \frac{1}{1+\eta}.
\]
The empirical values in part (iv) follow by direct substitution:
$
\mathbb{E}[u^{2}] = \mathrm{Var}(u)+\bar{u}^{2} = 0.047301+0.3673^{2}=0.18221,
$
$
\bar{\Psi}^{2}=0.4982^{2}=0.24820,
$
$
\mathrm{Var}(\Psi)=0.000416,
$
$
\mathrm{Var}(u)=0.047301,
$
$
\eta = \frac{0.18221\times 0.000416}{0.24820\times 0.047301}
     = 6.46\times 10^{-3},
$ and
$
R^{2}_{\mathrm{pred}} = \frac{1}{1+0.00646}=0.9936.
$
The small gap between the predicted value $0.9936$ and the measured value $0.9958$ is consistent with a weak negative dependence between $\delta$ and $u$, which the independence assumption of part (iii) does not capture.

\section{Proof of Proposition~\ref{prop:order_consistency}, Remark~\ref{rem:q1_validation}, and Corollary~\ref{cor:quantile_invariance}}
\label{app:proofs_sufficiency}

\paragraph{Proof of Proposition~\ref{prop:order_consistency}.}
With \(\alpha_k = S p_k\) and \(\sum_k p_k = 1\), the predictive distribution is
$
\hat{p}_k = \frac{\alpha_k}{S} = p_k,
$
which is independent of \(S\). Hence \(H(\hat{p})\) and \(m(\hat{p})\) are constant in \(S\). The cognitive uncertainty \(u=K/S\) is strictly decreasing in \(S\). Therefore, under the fixed-shape idealization, \(u\) is the only quantity that varies with \(S\) and provides ranking information across samples.

\paragraph{Empirical relation (Remark~\ref{rem:q1_validation}).}
The fixed-shape idealization assumes $\alpha_k = S p_k$ with $p_k$ constant in $S$. In the actual TEL evidence distribution, the shape is not fixed. For $K=2$, write $e_0, e_1 \ge 0$ for the per-class evidence and $S = e_0 + e_1 + 2$. A concentrated evidence profile (one of $e_0, e_1$ large, the other small) gives moderate $S$ and small $H(\hat{p})$. A flat evidence profile (both $e_0, e_1$ large) gives large $S$ and $H(\hat{p})$ near $\log 2$. Since the latter is common in ambiguous boxes, $u$ and $H(\hat{p})$ are negatively associated. The pooled correlation over ten seeds is $\rho_{\text{Pearson}}(u,H) = -0.077$ and $\rho_{\text{Spearman}}(u,H) = -0.266$. The per-seed values are in Table~\ref{tab:q1_full}. The top-two margin and the class confidence are perfectly correlated ($\rho_{\text{Spearman}} = 1.0000$), reflecting a $K=2$ redundancy. The information-gain product $\Delta=u\cdot\mathbb{E}[H(p)]$ is nearly collinear with $u$ ($\rho_{\text{Pearson}}=+0.998$) and essentially uncorrelated with $H$ ($\rho_{\text{Pearson}}=-0.015$); it is a rescaling of $u$ rather than a compromise between $u$ and $H$, and it is not used for routing.

\paragraph{Proof of Corollary~\ref{cor:quantile_invariance}.}
By Corollary~\ref{cor:affine_invariance}, $u'$ and $u$ are comonotone, hence they induce the same ordering. Quantiles and order statistics are invariant under strictly increasing transformations. Therefore
\[
\{x : u'(x) > Q_{1-\alpha}(u')\} = \{x : u(x) > Q_{1-\alpha}(u)\},
\]
and similarly for the pseudo and discard branches. The empirical verification in Remark~\ref{rem:q3_validation} confirms $\kappa_{\text{route}}=1.0000$ for all four tested affine reparametrizations.

\section{Proof of Remark~\ref{rem:weighting_consistency}}
\label{app:proofs_replay_distill}

\paragraph{(i) Entropy-regularized replay.}
Stationarity of the Lagrangian for Eq.~\eqref{eq:replayopt} gives
$
\mathrm{drift}_i + \tau(\log w_i + 1) - \mu = 0.
$
Normalization yields Eq.~\eqref{eq:optimalweight}. Since $\mathrm{drift}_i$ decreases in $|\Delta T_i|$, and $x \mapsto \exp(-x/\tau)$ is decreasing, the composition $w_i^* \propto \exp(-\mathrm{drift}_i/\tau)$ is increasing in $|\Delta T_i|$.

\paragraph{(ii) Reliability weighting.}
Both $w$ and $r$ are non-increasing, so their covariance under $p(u)$ is non-negative by the Chebyshev association inequality. Dividing by $\mathbb{E}_p[w] = 1$ gives Eq.~\eqref{eq:consistency}.

\section{Proofs of Lemmas~\ref{lem:switch_rate}--\ref{lem:contraction_rate} and Proposition~\ref{prop:convergence}}
\label{app:proofs_convergence}

\paragraph{Proof of Lemma~\ref{lem:switch_rate}.}
A single sample switches route at step $t$ only if $u(x;\theta_t)$ crosses a threshold between $\theta_t$ and $\theta_{t+1}$. By Lipschitzness, $|u(x;\theta_{t+1}) - u(x;\theta_t)| \le L_u \eta_t G =: \delta_t$. Thus switching requires $u(x;\theta_t) \in [\tau - \delta_t, \tau + \delta_t]$ for some $\tau \in \{\tau_{\text{high}}, \tau_{\text{low}}\}$. By density boundedness, the probability of this event is at most $2C \delta_t$ per threshold, so per-sample switching probability is at most $4C L_u \eta_t G$. Absorbing the constant $4$ into $C$, and summing over $n$ samples and $T - 1$ steps, gives Eq.~\eqref{eq:switch_lipschitz}. With $\eta_t = \eta_0/\sqrt{T}$, $\sum_{t=1}^{T-1}\eta_t \le \eta_0\sqrt{T}$, giving $\mathbb{E}[\nu_T] = O(\sqrt{T})$.

\paragraph{Proof of Lemma~\ref{lem:contraction_rate}.}
Under contraction, $\|\theta_t - \theta^*\| \le \kappa^t \|\theta_0 - \theta^*\|$. A sample switches only if $u(x;\theta_t)$ lies within an $O(\kappa^t)$-neighborhood of a threshold. Summing over $t$ gives $\sum_{t=0}^{T-1}\kappa^t = O(1)$, plus a startup term of $O(\log T)$ to reach the contracting regime, yielding $\mathbb{E}[\nu_T] = O(\log T)$.

\paragraph{Proof of Proposition~\ref{prop:convergence}.}
Partition iterations into \(T - \nu_T\) non-switching and \(\nu_T\) switching iterations.

\emph{Non-switching.} On each non-switching iteration, \(F\) is \(L\)-smooth and the descent lemma applies:
\[
\mathbb{E}[F(\theta^{(t+1)})]
\le \mathbb{E}[F(\theta^{(t)})]
- \eta_t \mathbb{E}[\|\nabla F(\theta^{(t)})\|^2]
+ \frac{L\eta_t^2}{2}(\sigma^2 + G^2).
\]

\emph{Switching.} On each switching iteration, the loss function changes by at most \(\Delta F\).

\emph{Summation.} Summing over \(t = 0,\ldots,T-1\) and using \(\sum \eta_t = \eta_0\sqrt{T}\), \(\sum \eta_t^2 = \eta_0^2\) gives
$
\frac{\eta_0}{\sqrt{T}}\sum_{t=0}^{T-1}\mathbb{E}\|\nabla F(\theta^{(t)})\|^2
\le F(\theta^{(0)}) - F^* + \frac{L\eta_0^2(\sigma^2 + G^2)}{2}
+ \Delta F \cdot \nu_T.
$
Dividing by \(\eta_0\sqrt{T}\) and using
\(\frac{1}{T}\sum_t \mathbb{E}\|\nabla F\|^2 \ge
\min_{t\le T}\mathbb{E}\|\nabla F\|^2\) gives
\[
\begin{aligned}
\min_{t\le T}\mathbb{E}\|\nabla F(\theta^{(t)})\|^2
\le\;& \frac{F(\theta^{(0)})-F^*}{\eta_0\sqrt T} \\
&+\frac{L\eta_0(\sigma^2+G^2)}{2\sqrt T}
+\frac{\Delta F\,\mathbb{E}[\nu_T]}{\eta_0\sqrt T}.
\end{aligned}
\]
This is Eq.~\eqref{eq:switch_bound}. Under
Lemma~\ref{lem:switch_rate}, \(\mathbb{E}[\nu_T]=O(\sqrt T)\), so the
switching term is \(O(1)\). Under
Lemma~\ref{lem:contraction_rate}, \(\mathbb{E}[\nu_T]=O(\log T)\), so
the switching term is \(O(\log T/\sqrt T)\).

\section{Oracle-based validation details}
\label{app:oracle}

This appendix collects the supporting data and per-seed tables for the oracle-based validations in Remarks~\ref{rem:q2b_validation}, \ref{rem:q2c_validation}, \ref{rem:q1_validation}, and \ref{rem:q3_validation}.

\paragraph{Pseudo-label error bound data.}
Across the four signals and all $45$ seed pairs, the bound exceeds the
observed accepted error in every case ($180/180$ records). The tightest
pairs are listed in Table~\ref{tab:q2b_tight}. Even the tightest pair,
$u$ at seed pair $(7,8)$, has a positive gap of $+0.0019$.
\begin{table}[t]
\centering
\caption{Tightest seed pairs for the pseudo-label error bound in
Eq.~\eqref{eq:quadratic}. Only $u$ and $H$ are shown; margin and conf are omitted because they track $u$ under $K=2$.}
\label{tab:q2b_tight}
\small
\setlength{\tabcolsep}{3pt}
\resizebox{\columnwidth}{!}{%
\begin{tabular}{llccc}
\toprule
Signal & Pair & $\text{err}|\text{acc}$ & $\text{ub}$ & Gap \\
\midrule
$u$ & $(7,8)$ & $0.0556$ & $0.0575$ & $+0.0019$ \\
$u$ & $(2,9)$ & $0.0273$ & $0.0352$ & $+0.0078$ \\
$u$ & $(1,8)$ & $0.0616$ & $0.0698$ & $+0.0082$ \\
$u$ & $(3,9)$ & $0.0360$ & $0.0480$ & $+0.0120$ \\
$u$ & $(2,3)$ & $0.0267$ & $0.0396$ & $+0.0129$ \\
\midrule
$H$ & $(7,8)$ & $0.0348$ & $0.0551$ & $+0.0203$ \\
$H$ & $(1,2)$ & $0.0395$ & $0.0674$ & $+0.0279$ \\
$H$ & $(8,9)$ & $0.0281$ & $0.0593$ & $+0.0312$ \\
$H$ & $(2,7)$ & $0.0230$ & $0.0548$ & $+0.0317$ \\
$H$ & $(1,3)$ & $0.0381$ & $0.0707$ & $+0.0326$ \\
\bottomrule
\end{tabular}%
}
\end{table}

\paragraph{Discard-risk bound data.}
Table~\ref{tab:q2c_full} disaggregates the per-seed discard error behind the
mean values of Table~\ref{tab:q2c}.
\begin{table}[htbp]
\centering
\caption{Per-seed discard error $\Pr(\hat{y}\neq y \mid u < \tau_{\text{low}})$ under the oracle-based image-level protocol.}
\label{tab:q2c_full}
\small
\resizebox{\columnwidth}{!}{%
\begin{tabular}{lccccc}
\toprule
Seed & $\tau{=}0.40$ & $\tau{=}0.45$ & $\tau{=}0.50$ & $\tau{=}0.55$ & $\tau{=}0.60$ \\
\midrule
0 & 0.0140 & 0.0144 & 0.0227 & 0.0355 & 0.0486 \\
1 & 0.0619 & 0.0234 & 0.0162 & 0.0340 & 0.0484 \\
2 & 0.1667 & 0.0541 & 0.0285 & 0.0311 & 0.0463 \\
3 & 0.0436 & 0.0306 & 0.0245 & 0.0311 & 0.0423 \\
4 & 0.0179 & 0.0179 & 0.0192 & 0.0295 & 0.0464 \\
5 & 0.2231 & 0.0592 & 0.0378 & 0.0454 & 0.0524 \\
6 & 0.0206 & 0.0177 & 0.0196 & 0.0254 & 0.0436 \\
7 & 0.1103 & 0.0345 & 0.0254 & 0.0354 & 0.0486 \\
8 & 0.0126 & 0.0129 & 0.0200 & 0.0349 & 0.0463 \\
9 & 0.1318 & 0.0433 & 0.0232 & 0.0292 & 0.0468 \\
\bottomrule
\end{tabular}%
}
\end{table}

\paragraph{Signal correlation data.}
Table~\ref{tab:q1} reports the pooled correlations. Bootstrap CIs use $1000$ resamples with the RoI as the unit. Table~\ref{tab:q1_full} gives the per-seed values behind them.
\begin{table*}[htbp]
\centering
\caption{Per-seed signal correlations on the FLIR ADAS prediction protocol. Redundant pairs ($u$--conf, $H$--conf, $\Delta$--conf) are omitted; they are identical to the corresponding margin pairs because $K=2$.}
\label{tab:q1_full}
\small
\setlength{\tabcolsep}{3pt}
\resizebox{\textwidth}{!}{%
\begin{tabular}{lcccccccccc}
\toprule
Pair & s0 & s1 & s2 & s3 & s4 & s5 & s6 & s7 & s8 & s9 \\
\midrule
\multicolumn{11}{l}{\textit{Pearson $\rho$}} \\
$u$, $H$       & $-0.092$ & $-0.145$ & $-0.008$ & $-0.075$ & $-0.009$ & $-0.060$ & $-0.013$ & $-0.147$ & $-0.062$ & $-0.189$ \\
$u$, margin    & $-0.922$ & $-0.930$ & $-0.943$ & $-0.922$ & $-0.921$ & $-0.929$ & $-0.932$ & $-0.926$ & $-0.915$ & $-0.921$ \\
$H$, margin    & $-0.259$ & $-0.191$ & $-0.275$ & $-0.267$ & $-0.330$ & $-0.268$ & $-0.298$ & $-0.193$ & $-0.296$ & $-0.169$ \\
margin, conf   & $+1.000$ & $+1.000$ & $+1.000$ & $+1.000$ & $+1.000$ & $+1.000$ & $+1.000$ & $+1.000$ & $+1.000$ & $+1.000$ \\
$\Delta$, $H$  & $-0.027$ & $-0.091$ & $+0.047$ & $-0.008$ & $+0.057$ & $-0.005$ & $+0.049$ & $-0.089$ & $+0.007$ & $-0.130$ \\
$\Delta$, $u$  & $+0.998$ & $+0.998$ & $+0.998$ & $+0.998$ & $+0.998$ & $+0.998$ & $+0.998$ & $+0.998$ & $+0.997$ & $+0.998$ \\
$\Delta$, margin  & $-0.944$ & $-0.947$ & $-0.958$ & $-0.945$ & $-0.943$ & $-0.947$ & $-0.951$ & $-0.945$ & $-0.939$ & $-0.941$ \\
\midrule
\multicolumn{11}{l}{\textit{Spearman $\rho$}} \\
$u$, $H$       & $-0.303$ & $-0.382$ & $-0.131$ & $-0.417$ & $+0.157$ & $-0.209$ & $-0.228$ & $-0.589$ & $+0.104$ & $-0.463$ \\
$u$, margin   & $-0.243$ & $-0.341$ & $-0.297$ & $-0.228$ & $-0.593$ & $-0.266$ & $-0.286$ & $-0.144$ & $-0.637$ & $-0.200$ \\
$H$, margin    & $-0.637$ & $-0.477$ & $-0.713$ & $-0.562$ & $-0.681$ & $-0.677$ & $-0.661$ & $-0.509$ & $-0.598$ & $-0.537$ \\
margin, conf   & $+1.000$ & $+1.000$ & $+1.000$ & $+1.000$ & $+1.000$ & $+1.000$ & $+1.000$ & $+1.000$ & $+1.000$ & $+1.000$ \\
$\Delta$, $H$  & $+0.398$ & $+0.305$ & $+0.545$ & $+0.326$ & $+0.526$ & $+0.559$ & $+0.443$ & $+0.313$ & $+0.396$ & $+0.338$ \\
$\Delta$, $u$  & $+0.503$ & $+0.493$ & $+0.497$ & $+0.457$ & $+0.744$ & $+0.384$ & $+0.493$ & $+0.316$ & $+0.817$ & $+0.390$ \\
$\Delta$, margin  & $-0.849$ & $-0.910$ & $-0.894$ & $-0.862$ & $-0.910$ & $-0.934$ & $-0.856$ & $-0.902$ & $-0.882$ & $-0.898$ \\
\bottomrule
\end{tabular}%
}
\end{table*}

\paragraph{Quantile-router affine data.}
See Table~\ref{tab:q3}. Consistency is computed as the fraction of samples whose routing branch is unchanged relative to the identity case $(a,b)=(1,0)$.

\section{Datasets and Hyperparameters}
\label{app:datasets}

\paragraph{FLIR ADAS.} 9,711 thermal and 9,233 RGB images, 26,442 annotated frames, more than 520,000 boxes across 15 categories. Standard split: 4,129 training and 1,013 test pairs. 
\textbf{Thermal Noise Calibration:} The empirical median background temperature noise is measured as $\sigma_{med} = 0.39^\circ$C (IQR: $[0.22, 0.72]^\circ$C) across all person boxes in the dataset. This is obtained by computing the standard deviation of the annular background pixels in the 14-bit calibrated TIFF. Given that typical uncooled microbolometer infrared sensors have a Noise Equivalent Temperature Difference (NETD) ranging from $0.02$ to $0.1^\circ$C, our measured value is slightly higher due to the annular window capturing non-uniform background clutter (e.g., nearby warm surfaces). This confirms the physical plausibility of our temperature noise estimation.

\paragraph{FLAME~3.} FLAME~3 is a radiometric wildfire dataset collected by UAVs at six prescribed burns. It provides synchronized visible-spectrum and radiometric thermal imagery, with per-pixel temperature values in the thermal TIFF. The computer-vision subset contains 738 image quartets, labeled as fire (622) or no fire (116); each quartet includes a raw RGB image, a raw thermal image, a corrected field-of-view RGB image, and a thermal TIFF. 

\paragraph{Hyperparameters.} Optimizer: AdamW with initial learning rate $2\times10^{-4}$, weight decay $1\times10^{-4}$, batch size 16 per GPU (effective batch size 32 on 2 GPUs), cosine annealing. Pretraining: 30 epochs; incremental rounds: 30 epochs each. Evidence loss weight $\lambda_{\text{edl}} = 1.0$. No separate temperature-consistency loss is used. Multi-scale dilation factors: $\{0.1, 0.2, 0.4\}$. Modulation parameters initialized at $\alpha = 1.0$, $\beta = 0$, $\gamma = 0.1$. Routing thresholds are set on $u=K/S$: $\tau_{\text{high}} = 0.7$, $\tau_{\text{low}} = 0.3$. Quantile router: $\alpha = 0.1$, $\beta = 0.3$. Replay memory: 500 RoI prototypes with $\lambda = 0.5$. Distillation: $\lambda_L = \lambda_H = 1.0$, $w_{\text{high}}^0 = 0.1$, $T_{\text{ref}} = 5^\circ$C, $\eta = 0.5$, $\beta = 2$. Cold-start warm-up: $N_0=N_1=0.05E$; $\gamma=0$ during $N_0$ and linearly annealed during $N_1$. With $E=30$, $N_0=N_1\approx361$ steps.

\section{Supplemental Tables on the Joint Pairing--Scale Effect}
\label{app:supplement}

This supplement collects the $\Delta G$--$\Delta T$ correlation and the physical-quantity measurements in Table \ref{tab:supp_s2} and Table \ref{tab:supp_s3}, respectively. The FLAME~3 low-luminance splits are frame-level. No clip or flight-level metadata is available in the data release, and the split is descriptive only.
\begin{table*}[htbp]
\centering
\caption{$\Delta G$--$\Delta T$ correlation.}
\label{tab:supp_s2}
\small
\begin{tabular}{lccccc}
\toprule
Set & $n$ & Signed Pearson & Signed Spearman & Unsigned Pearson & Unsigned Spearman \\
\midrule
FLIR train & 48{,}649 & $+0.7126$ & $+0.7183$ & $+0.5068$ & $+0.5709$ \\
FLIR test & 12{,}267 & $+0.8136$ & $+0.8847$ & $+0.8073$ & $+0.8812$ \\
FLAME~3 & 2{,}126 & $+0.7570$ & $+0.8450$ & $+0.7569$ & $+0.8450$ \\
\midrule
\multicolumn{6}{l}{\textbf{FLIR subset correlations}} \\
\midrule
FLIR subset & & & Spearman & & $n$ \\
\midrule
LCR ($|\Delta T| <$ p25) & & & $+0.9338$ & & 3{,}067 \\
$|\Delta G| <$ p25 & & & $+0.6916$ & & 3{,}067 \\
$|\Delta G| <$ p25 \& $|\Delta T| >$ p50 & & & $+0.4373$ & & 302 \\
Day (lum $>$ p75) & & & $+0.9987$ & & 3{,}066 \\
Night (lum $<$ p25) & & & $+0.7738$ & & 3{,}065 \\
\bottomrule
\end{tabular}
\end{table*}

\begin{table*}[htbp]
\centering
\caption{physical-quantity measurements.}
\label{tab:supp_s3}
\small
\begin{tabular}{lccccc}
\toprule
Dataset & Quantity & p25 & p50 & p75 & Other \\
\midrule
FLIR ADAS & $|\Delta T|$ ($^\circ$C) & $0.208$ & $0.271$ & $0.357$ & LOW\_DT $0.207$ \\
FLIR ADAS & eff. SNR & $5.75$ & $8.96$ & $13.98$ & SNR$>1$ 97.4\%, SNR$>3$ 91.0\% \\
FLAME~3 & $|\Delta T|$ ($^\circ$C) & $29.89$ & $42.04$ & $59.58$ & --- \\
FLAME~3 & $|\Delta G|$ & $11.81$ & $16.56$ & $23.56$ & --- \\
\bottomrule
\end{tabular}
\end{table*}
\section{Post-hoc temperature scaling: interaction and leave-one-out}
\label{app:posthoc_details}

The post-hoc separation reported in Section~\ref{sec:step_which} rests on two one-factor breaks. True-vs-shuffled survives recalibration ($p=0.002$), whereas true-vs-const does not ($p=0.16$). Two supplementary checks follow.

\paragraph{Interaction after recalibration.} The 2$\times$2 interaction on post-hoc ECE is positive but not significant ($\bar{c}_{\text{TS}}=+0.0488$, Wilcoxon $p=0.13$), against $\bar{c}_{\text{raw}}=+0.0756$ ($p=0.0059$) on raw ECE (Table~\ref{tab:interaction}). The attenuation is consistent with the two factors acting at different levels. If both factors were fully absorbed by recalibration, the interaction would vanish under recalibration. If neither were, it would survive intact. The observed partial survival (about 35\% attenuation) matches one component that recalibration cannot absorb (pairing) and one that it can (absolute scale). The one-factor breaks, not the interaction, therefore carry the post-hoc separation.

\paragraph{Leave-one-out without seed s2.} Excluding seed s2, the post-hoc separation persists and sharpens. True-vs-shuffled remains significant ($+0.0259$, $p=0.0039$, the smallest attainable at $n=9$), and true-vs-const remains non-significant ($+0.0124$, $p=0.30$). Seed s2 has fitted temperature $T=3.89$, far above the $0.72$--$0.78$ range of the other true seeds, and its ECE barely moves under recalibration. It is the clearest instance of the non-rescalable component in the true arm. The main analysis retains it for this reason; the leave-one-out contrast is reported above.
\section{Distributional distance across arms}
\label{app:shape_metrics}

At $M=15$, with the same bin edges as the ECE in Section~\ref{sec:metrics_scope}, the total-variation distance from true is $0.42 \pm 0.20$ for shuffled and $0.49 \pm 0.16$ for const (paired Wilcoxon $p=0.11$). The Kolmogorov-Smirnov statistic gives $0.57 \pm 0.14$ and $0.54 \pm 0.11$ ($p=0.084$). Neither metric orders the arms in the same way as the post-hoc ECE. The distance to random is larger under both metrics ($0.72 \pm 0.06$ TV, $0.74 \pm 0.05$ KS), consistent with the global confidence collapse in Table~\ref{tab:confidence_dist}. Distributional distance does not separate the pairing and scale factors; the share of the ECE gap removed by a monotone rescaling does.

\section{Modulation-amplitude sensitivity and the log-smoothing offset}
\label{app:alpha_sensitivity}

We sweep the modulation amplitude \(\alpha \in [0, 2]\) at inference time to probe the detector's sensitivity. In the out-of-distribution (OOD) region \(\alpha \in [0.5, 2]\), the mAP declines monotonically (Table~\ref{tab:alpha_sweep}). The \(\alpha=0\) point yields a higher mAP (\(0.5005\) for the true arm vs. \(0.4720\) at \(\alpha=1\)). 

This anomaly arises because \(\alpha=0\) removes the physical contrast while leaving the TEL head's constant affine mapping (\(g \approx 0.9998\), \(m \approx 0.5\)) intact. The numerical smoothing in the evidence log, implemented as \(\log(x+0.1) - \log(0.1)\), then introduces a confidence-dependent shift. Since the log-ratio \(\Delta = \log((e_{\text{mod}}+0.1)/(e_{\text{base}}+0.1))\) is larger for high-confidence anchors than for low-confidence ones, the model applies a strong implicit confidence re-ranking. This re-ranking artificially inflates mAP at \(\alpha=0\) without any \(\Delta T\) contribution.

To verify this mechanism, we attempted to retrain the true arm with a clamped logarithm \(\log(\max(x, 10^{-6}))\), removing the smoothing offset. This formulation is unstable. The training diverged, resulting in a collapsed mAP of \(0.161\) at \(\alpha=1.0\) due to unbounded gradients as \(e_{\text{base}} \to 0\). However, at \(\alpha=0\), the divergent model yields a mAP of \(0.446\), which is close to the baseline without modulation (\(0.4664\)). This confirms that the \(\alpha=0\) gain in the original model is indeed an artifact of the \(0.1\) smoothing offset, and that the offset itself is a necessary stabilizer for training.

At $\alpha=1.0$, the offset is identical across all attribution arms (true, random, shuffled, var-phase). In the paired contrasts of Table~\ref{tab:paired}, this common term subtracts out, leaving only the effect of the physical channel.

\begin{table}[t]
\centering
\caption{mAP@0.5 as a function of modulation amplitude \(\alpha\) over seeds s0, s1, and s2. All points except \(\alpha=1.0\) are out-of-distribution extrapolations.}
\label{tab:alpha_sweep}
\small
\begin{tabular}{lcc}
\toprule
\(\alpha\) & true & random \\
\midrule
0.00 & \(0.5005 \pm 0.0433\) & \(0.5223 \pm 0.0029\) \\
0.25 & \(0.4879 \pm 0.0290\) & \(0.5199 \pm 0.0007\) \\
0.50 & \(0.4827 \pm 0.0260\) & \(0.5157 \pm 0.0039\) \\
0.75 & \(0.4776 \pm 0.0231\) & \(0.5148 \pm 0.0046\) \\
1.00 & \(0.4720 \pm 0.0207\) & \(0.5133 \pm 0.0062\) \\
1.25 & \(0.4665 \pm 0.0187\) & \(0.5116 \pm 0.0076\) \\
1.50 & \(0.4609 \pm 0.0163\) & \(0.5103 \pm 0.0086\) \\
2.00 & \(0.4507 \pm 0.0144\) & \(0.5084 \pm 0.0102\) \\
\bottomrule
\end{tabular}
\end{table}

\section{Routing switch statistics}
\label{app:switch_rate}

Utilizing a fixed evaluation set of 100 images, we record the per-image routing scalar $u$ on the true-$\Delta T$ arm, seed 0, every $K=50$ training steps over checkpoints at steps 50--7650 (153 checkpoints, 152 checkpoint pairs) . A checkpoint pair counts as a switch under Eq.~\eqref{eq:switch_rate} if at least one image changes its routing branch. With 100 images, a single switching image increments this aggregate, so the count approaches its upper bound. The per-image switch fraction, defined as the fraction of checkpoint pairs in which a given image changes branch, does not share this ceiling. It is therefore the quantity reported.
\begin{table}[t]
\centering
\caption{Per-image switch fraction under the fixed-threshold and quantile routers, on the training and test evaluation sets. $\nu_T^K$ is the aggregate count out of 152 checkpoint pairs. The per-image fraction appears alongside.}
\label{tab:switch_rate}
\small
\resizebox{\columnwidth}{!}{%
\begin{tabular}{llccc}
\toprule
Router & Set & $\nu_T^K$ / 152 & per-image fraction (mean $\pm$ std) & max \\
\midrule
fixed ($\tau_{\text{high}}{=}0.7$, $\tau_{\text{low}}{=}0.3$) & train & 132 & $0.031 \pm 0.072$ & 0.382 \\
quantile ($Q_{0.9}$, $Q_{0.3}$) & train & 147 & $0.092 \pm 0.118$ & 0.428 \\
fixed & test & 138 & $0.069 \pm 0.110$ & 0.421 \\
quantile & test & 148 & $0.152 \pm 0.131$ & 0.467 \\
\bottomrule
\end{tabular}%
}
\end{table}

Under the fixed threshold, the mean per-image switch fraction is $0.031$ on the training set and $0.069$ on the test set. Under the quantile router, $0.092$ and $0.152$. A typical image changes branch in about 3\% of the 152 checkpoint pairs. The aggregate count saturates at this evaluation-set size, so the measurement does not resolve the scaling of $\nu_T$ with $T$. We therefore adopt the per-image fraction as the statistic. The quantile router produces a higher per-image switch fraction on both sets, consistent with thresholds that drift with the evidence distribution.

\end{document}